\documentclass{article}

\usepackage{arxiv}

\usepackage[utf8]{inputenc} 
\usepackage[T1]{fontenc}    
\usepackage{hyperref}       
\usepackage{url}            
\usepackage{booktabs}       
\usepackage{amsfonts}       
\usepackage{nicefrac}       
\usepackage{microtype}      
\usepackage{lipsum}		
\usepackage{graphicx}
\usepackage[numbers,sort&compress]{natbib}
\usepackage{doi}
\usepackage{amsmath}

\title{\fontsize{15}{18}\selectfont
Accurate and Efficient Hybrid-Ensemble Atmospheric Data Assimilation in Latent Space with Uncertainty Quantification}

\author{
Hang Fan$^{1,2,*}$, Juan Nathaniel$^{1,2}$, Yi Xiao$^{3,5}$, Ce Bian$^{1,2}$,
Fenghua Ling$^{3}$,\\[2pt]
\textbf{Ben Fei$^{3,4,*}$, Lei Bai$^{3}$, Pierre Gentine$^{1,2}$}\\
\\
\small $^{1}$Department of Earth and Environmental Engineering, 
School of Engineering and Applied Sciences, \\
\small Climate School, Columbia University, New York, NY, USA\\[2pt]
\small $^{2}$Learning the Earth with Artificial Intelligence and Physics (LEAP) Center, Columbia University, New York, NY, USA\\[2pt]
\small $^{3}$Shanghai Artificial Intelligence Laboratory, Shanghai, China\\[2pt]
\small $^{4}$The Chinese University of Hong Kong, Hong Kong, China\\[2pt]
\small $^{5}$Department of Computer Science and Technology, 
Tsinghua University, Beijing, China\\
}

\date{}

\renewcommand{\shorttitle}{Hybrid-Ensemble Atmospheric Data Assimilation in Latent Space}

\hypersetup{
pdftitle={A template for the arxiv style},
pdfsubject={q-bio.NC, q-bio.QM},
pdfauthor={David S.~Hippocampus, Elias D.~Striatum},
pdfkeywords={First keyword, Second keyword, More},
}

\begin{document}
\maketitle

\begin{abstract}
Data assimilation (DA) combines model forecasts and observations to estimate the optimal state of the atmosphere with its uncertainty, providing initial conditions for weather prediction and reanalyses for climate research.
Yet, existing traditional and machine-learning DA methods struggle to achieve accuracy, efficiency and uncertainty quantification simultaneously.
Here, we propose HLOBA (Hybrid-Ensemble Latent Observation–Background Assimilation), a three-dimensional hybrid-ensemble DA method that operates in an atmospheric latent space learned via an autoencoder (AE).
HLOBA maps both model forecasts and observations into a shared latent space via the AE encoder and an end-to-end Observation-to-Latent-space mapping network (O2Lnet), respectively, and fuses them through a Bayesian update with weights inferred from time-lagged ensemble forecasts.
Both idealized and real-observation experiments demonstrate that HLOBA matches dynamically constrained four-dimensional DA methods in both analysis and forecast skill, while achieving end-to-end inference–level efficiency and theoretical flexibility applies to any forecasting model.
Moreover, by exploiting the error decorrelation property of latent variables, HLOBA enables element-wise uncertainty estimates for its latent analysis and propagates them to model space via the decoder.
Idealized experiments show that this uncertainty highlights large-error regions and captures their seasonal variability.
\end{abstract}

\newcommand\blfootnote[1]{%
  \begingroup
  \renewcommand\thefootnote{}%
  \footnotetext[0]{#1}%
  \endgroup
}

\blfootnote{$^\ast$Corresponding author. Email: hf2526@columbia.edu, benfei@cuhk.edu.hk}


\section*{Introduction}\label{sec1}
Data assimilation (DA) aims to estimate the optimal state of a dynamical system and its uncertainty (the posterior) by updating numerical model predictions (the prior) with observations according to their error statistics~\cite{carrassiDataAssimilationGeosciences2018, evensenDataAssimilationFundamentals2022}. 
As the backbone of modern atmospheric science, DA produces high-quality analyses (i.e., posterior optimal state estimates) that initialize numerical weather prediction and construct reanalysis datasets spanning decades of historical atmosphere that underpin climate research.
Furthermore, the uncertainty estimates associated with DA analyses enable probabilistic forecasting that is critical for extreme-event prediction~\cite{houtekamerReviewEnsembleKalman2016a}, support confidence assessment in climate studies~\cite{compoTwentiethCenturyReanalysis2011}, and guide the deployment of observing systems~\cite{tanObservingsystemImpactAssessment2007}.

Despite the remarkable success of traditional DA methods in producing accurate analyses, fundamental challenges remain. 
First, accurate uncertainty quantification for traditional DA analyses is difficult in practice. Operational DA systems typically run an Ensemble of Data Assimilations (EDA) to represent analysis uncertainty~\cite{isaksenEnsembleDataAssimilations2010a,houtekamerReviewEnsembleKalman2016a,bannisterReviewOperationalMethods2017a}, which requires both careful design of the ensemble members and measures to avoid ensemble collapse or divergence during DA cycling. However, even with considerable effort and computing resources, the resulting analysis uncertainty estimates inevitably exhibit substantial sampling noise and spurious patterns, as the ensemble sizes used in practice (typically on the order of $10^2$–$10^3$ members) are still far smaller than the model’s dimension (often exceeding $10^8$). Second, the continued increase in model resolution and observation volume is driving a rapid escalation in the computational and memory demands of DA, a burden that becomes even more severe when uncertainty quantification is required. These challenges highlight the need for new DA frameworks that jointly improve analysis accuracy, uncertainty quantification, and computational efficiency.

To meet these challenges, data-driven machine-learning (ML) approaches are emerging as a key direction for advancing DA~\cite{geerLearningEarthSystem2021, dubenMachineLearningECMWF2021,chengMachineLearningData2023a}.
One increasingly popular avenue is generative DA, which uses generative artificial intelligence (GenAI) to approximate the posterior distribution and draw analysis samples from it.
Such approaches can relax the Gaussian assumptions underlying many traditional DA methods~\cite{carrassiDataAssimilationGeosciences2018} and provide a more faithful analysis uncertainty estimation through extensive sampling.
Early implementations relied on generative adversarial networks (GANs)~\cite{baoCouplingEnsembleSmoother2020, baoVariationalAutoencoderGenerative2022, hamPartialconvolutionimplementedGenerativeAdversarial2024} but have recently shifted towards diffusion models~\cite{rozetScorebasedDataAssimilation, huangDiffDADiffusionModel2024,quDeepGenerativeData2024, sunAlignDAAlignScorebased2025,manshausenGenerativeDataAssimilation2025,sunLOSDALatentOptimization2025,savaryTrainingFreeDataAssimilation2025}, in which observations guide an iterative denoising process to produce analysis states. 
However, current generative DA has not shown clear improvements in analysis accuracy over traditional DA methods and remains computationally expensive, particularly for uncertainty quantification.

Another emerging avenue is latent data assimilation (LDA)~\cite{peyronLatentSpaceData2021, amendolaDataAssimilationLatent2021, chengMultidomainEncoderDecoder2024, melinc3DVarDataAssimilation2024, fanNovelLatentSpace2025, fanNovelLatentSpace2025a, fanPhysicallyConsistentGlobal2026,melincUnifiedNeuralBackgroundError2026}, which applies traditional DA methods under Gaussian distribution assumptions in a reduced-order latent space of the atmosphere learned with autoencoders (AEs). 
Many studies have shown that latent representations can encode complex spatial and multivariate dependencies among atmospheric variables, with associated error covariances that are often small enough to be neglected~\cite{chengMultidomainEncoderDecoder2024,melinc3DVarDataAssimilation2024,fanNovelLatentSpace2025a,fanPhysicallyConsistentGlobal2026,melincUnifiedNeuralBackgroundError2026}. These properties simplify uncertainty estimation in latent space, making LDA easier to implement and more effective than traditional DA methods applied directly in model space~\cite{fanPhysicallyConsistentGlobal2026}.
However, LDA is still nascent, and the use of latent representations for uncertainty estimation remains largely unexplored.
In addition, the state-of-the-art LDA method, latent four-dimensional variational data assimilation (L4DVar), relies on iterative variational optimization in a differentiable framework. In practice, this simplification entails (1) additional effort to implement both the forecast model and the observation operator within the same differentiable framework and (2) several-fold higher memory and computational cost than the forward inference of these operators alone, limiting scalability.

In parallel, some studies dispense with the inherently probabilistic treatment of DA and instead train ML models to learn an end-to-end mapping from observations and model forecasts (background fields) to reanalysis in a deterministic manner~\cite{chenEndtoendArtificialIntelligence2024,allenEndtoendDatadrivenWeather2025,xuFuXiDAGeneralizedDeep2025,xiangADAFArtificialIntelligence2025,sunDatatoforecastMachineLearning2025}.
This strategy can deliver high accuracy at low computational cost, demonstrating the efficacy of end-to-end learning in extracting information from observations. 
By bypassing explicit observation operators and the complex estimation of error statistics, this end-to-end strategy significantly reduces the implementation overhead, an advantage particularly pronounced for indirect observations, such as satellites.
However, this simplicity comes at the substantial expense of (1) the ability to quantify uncertainty in the analysis, (2) the flexibility to handle evolving background and observation uncertainty configurations that deviate from those encountered during training, and (3) the potential to surpass the reanalysis data used as truth for training, even when more observations are available.
Accordingly, these end-to-end approaches function primarily as powerful analysis models rather than as DA methods.

In this study, we bridge end-to-end learning with the probabilistic structure of LDA by introducing an additional neural network, the observation-to-latent mapping network (O2Lnet)~\cite{fanNovelLatentSpace2025, fanNovelLatentSpace2025a}, to map observations directly into the latent space that represents the global atmosphere. 
This design enables Bayesian assimilation of the encoded observations and background fields in a unified space, with their uncertainties estimated in a hybrid manner that blends a fixed climatological component with a flow-dependent ensemble component~\cite{hamillHybridEnsembleKalman2000a,lorencPotentialEnsembleKalman2003a,claytonOperationalImplementationHybrid2013}.
We refer to this method as \textbf{HLOBA} (\textbf{H}ybrid-Ensemble \textbf{L}atent \textbf{O}bservation–\textbf{B}ackground \textbf{A}ssimilation).
Despite operating as a three-dimensional DA scheme, HLOBA delivers assimilation and forecast skills comparable to an ensemble implementation of dynamically constrained L4DVar, while requiring only about 3\% of the runtime and 20\% of the memory. 
Moreover, HLOBA can estimate analysis errors even with a small ensemble size. We demonstrate that these advantages primarily arise from the synergy between the strong representational capacity of end-to-end mapping and the associated ensemble-based uncertainty estimation.

\section*{Results}\label{sec2}

\subsection*{Description of HLOBA method}
\begin{figure}[ht]
    \centering
    \includegraphics[width=\linewidth]{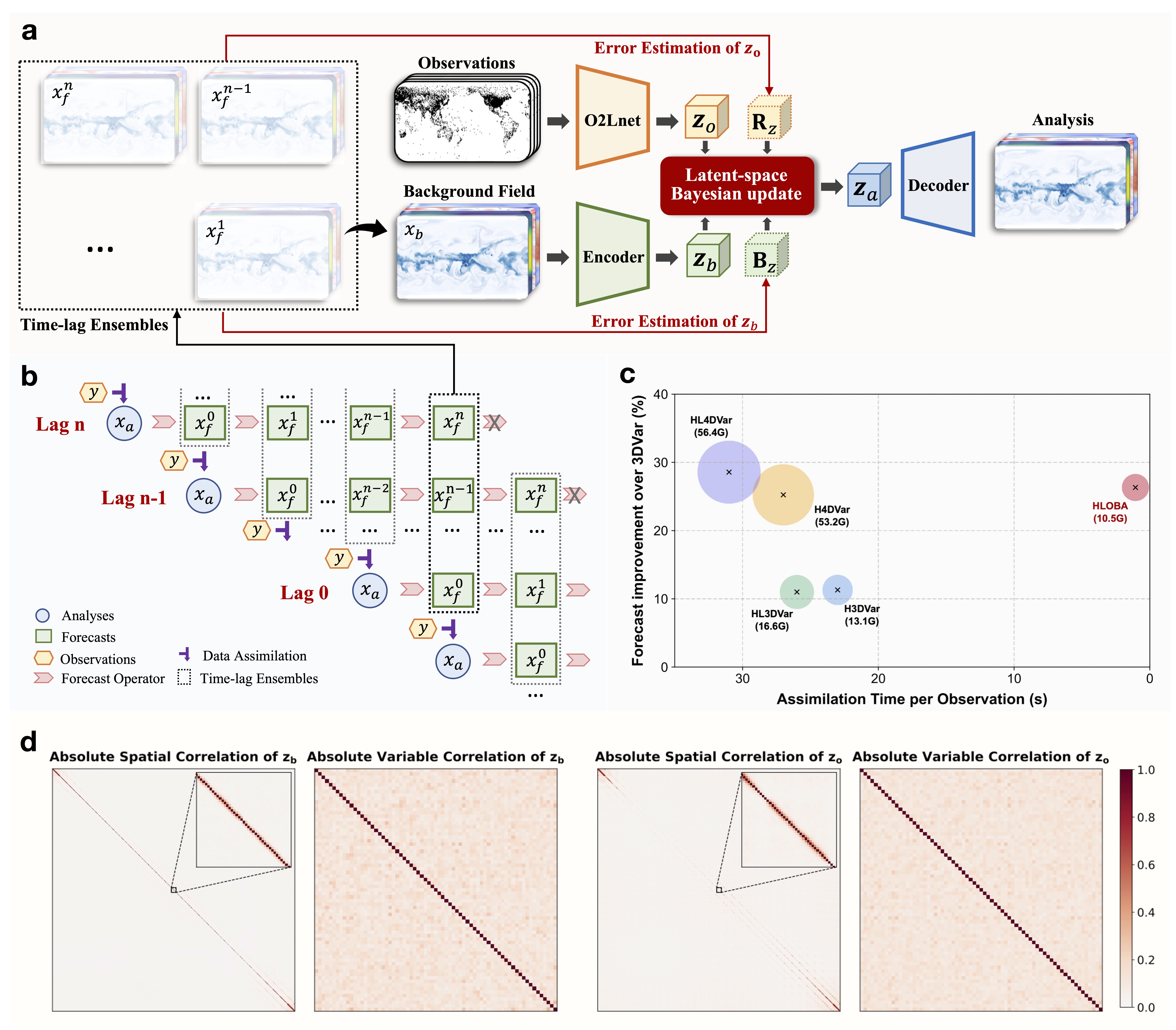}
    \caption{\textbf{Overview of the HLOBA method.} 
    \textbf{a}, HLOBA pipeline. The background field and observations are mapped into a shared latent space using the Encoder and O2Lnet, respectively, and fused in a Bayesian manner based on their uncertainty estimates derived from time-lagged ensembles. The resulting latent analysis is then decoded to obtain the final analysis in model space.
    \textbf{b}, Generation of time-lagged ensembles. Analyses produced at different assimilation cycles are propagated forward to form a flow-dependent ensemble that represents background uncertainty without requiring parallel ensemble forecasts.
    \textbf{c}, Comparison of HLOBA with other hybrid DA methods when assimilating real observations on an NVIDIA A100. The x-axis shows the assimilation time per observation, and the y-axis shows forecast improvement relative to 3DVar. Marker size indicates GPU memory usage.
    \textbf{d}, Climatological estimates of error correlations for latent-space background $\boldsymbol{z}_b$ and observation $\boldsymbol{z}_o$ (see Methods for details). Inter-variable correlations are computed across latent dimensions at each grid point and averaged in absolute value; spatial correlations are computed across grid points within each latent dimension and averaged in absolute value.
    }
    \label{fig:fig1}
\end{figure}

HLOBA maps both observations and background fields into a shared latent space for Bayesian assimilation, which relies on three neural network modules: an encoder, a decoder, and an observation-to-latent mapping network (O2Lnet) (Figure~\ref{fig:fig1}a).
The encoder and decoder are trained jointly as an AE to define a latent representation of the high-dimensional global atmosphere, with the encoder compressing atmospheric model states into a low-dimensional latent space and the decoder reconstructing them back to model space. 
O2Lnet is then trained using encoded reanalysis fields as targets and observations sampled from the same fields as inputs, thereby learning an end-to-end mapping from observations to the latent space.
Through the encoder and O2Lnet, both the background state and the observations are represented in a unified latent space. 
In this space, by assuming the errors follow a zero-mean Gaussian distribution, the posterior mean (i.e., the latent analysis) and the associated uncertainty are obtained by optimally combining the latent background state and latent observations, weighted by their respective error covariance matrices $\mathbf{B}_z$ and $\mathbf{R}_z$. 
Finally, the analysis and its uncertainty in model space are recovered by decoding the latent analysis and its associated uncertainty.
We note that this AE-O2Lnet framework can be trained without any real observations, as it relies solely on reanalysis fields for self-supervision, yet can be generalized to assimilate real observations.
Further details are provided in the Methods section.

We train both the AE and O2Lnet on the ERA5 reanalysis dataset~\cite{carrassiDataAssimilationGeosciences2018} and evaluate the potential of this LDA framework for global atmospheric analyses and forecasts. 
The ERA5 fields are regridded to a $1.41^\circ$ horizontal resolution to reduce computational cost and comprise four surface variables and five upper-air variables on 13 vertical levels, yielding a model state of size $69 \times 256 \times 128$ (variables$\times$longitude$\times$latitude). The latent space has dimension $69 \times 64 \times 32$ with the same variables–zonal-meridional structure as the model state, corresponding to a compression ratio of 16. 
As a proof-of-concept study, we assimilate only surface and radiosonde observations from the 2017 GDAS (Global Data Assimilation System), whose simple observation operators make it easy to generate simulated observations for training O2Lnet.

Many existing LDA approaches find that $\mathbf{B}_z$ is approximately diagonal~\cite{melinc3DVarDataAssimilation2024, melincUnifiedNeuralBackgroundError2026, fanNovelLatentSpace2025a, fanPhysicallyConsistentGlobal2026}, and we find that this property also holds for $\mathbf{R}_z$ (Figure~\ref{fig:fig1}d). This structure justifies neglecting cross-covariances among latent variables, allowing $\mathbf{B}_z$ and $\mathbf{R}_z$ to be approximated as diagonal matrices. HLOBA can therefore perform assimilation and uncertainty estimation independently for each latent variable. This property offers a major computational advantage and markedly simplifies uncertainty quantification, since a diagonal covariance matrix can be estimated with only two ensemble members.

We conduct cycling assimilation and forecast experiments using a deterministic ML-based forecast model with a 6-hour time step. 
Since deterministic ML forecast models are often insensitive to initial-condition perturbations~\cite{selzCanArtificialIntelligenceBased2023}, yielding poorly spread perturbation-based ensembles, we estimate $\mathbf{B}_z$ and $\mathbf{R}_z$ from a time-lagged ensemble formed by forecasts of different lead times that verify at the same analysis time (Figure~\ref{fig:fig1}b)~\cite{houtekamerSystemSimulationApproach1996,lorencImprovingEnsembleCovariances2017,caronImprovingBackgroundError2019a}. For example, at analysis time $t$, the ensemble is constructed from forecasts that are all valid at $t$ but are initialized from the analyses at $t-6\,\mathrm{h}$, $t-12\,\mathrm{h}$, $t-18\,\mathrm{h}$, and earlier cycles.
However, a main limitation of the time-lagged strategy is that ensemble quality deteriorates as members with increasingly long lead times are added. We therefore restrict the ensemble size to 3, 6 and 9 members to assess the sensitivity of our schemes on ensemble size. 
To mitigate sampling noise and spurious correlations associated with small ensembles, we use a hybrid formulation~\cite{hamillHybridEnsembleKalman2000a,lorencPotentialEnsembleKalman2003a,claytonOperationalImplementationHybrid2013} that blends ensemble-derived error covariances with a static climatological component estimated from extensive historical data.

\subsection*{Performance and efficiency of HLOBA}
We compare HLOBA against a set of hybrid DA schemes that use time-lagged ensembles, including hybrid three-dimensional variational assimilation (H3DVar)~\cite{hamillHybridEnsembleKalman2000a}, hybrid four-dimensional variational assimilation (H4DVar)~\cite{claytonOperationalImplementationHybrid2013}, and their latent-space counterparts (HL3DVar and HL4DVar). Further implementation details are given in the Methods section. 
To ensure robustness, we perform year-long cycling assimilation–forecast experiments for 2017 under both idealized and real-observation settings. Each DA cycle assimilates observations from four consecutive time steps and then issues a 5-day forecast used to compute forecast errors. Three-dimensional DA methods (3D-DA: H3DVar, HL3DVar, and HLOBA) assimilate the four observation steps sequentially, whereas four-dimensional DA methods (4D-DA: H4DVar and HL4DVar), in which the model dynamics constrains the analysis, assimilate them jointly. For each scheme, we determine the optimal ensemble size (3, 6, or 9 members) and associated hyperparameters using experiments conducted in January, and then apply the resulting configuration to the full-year simulations. 

\begin{figure}[ht]
    \centering
    \includegraphics[width=\linewidth]{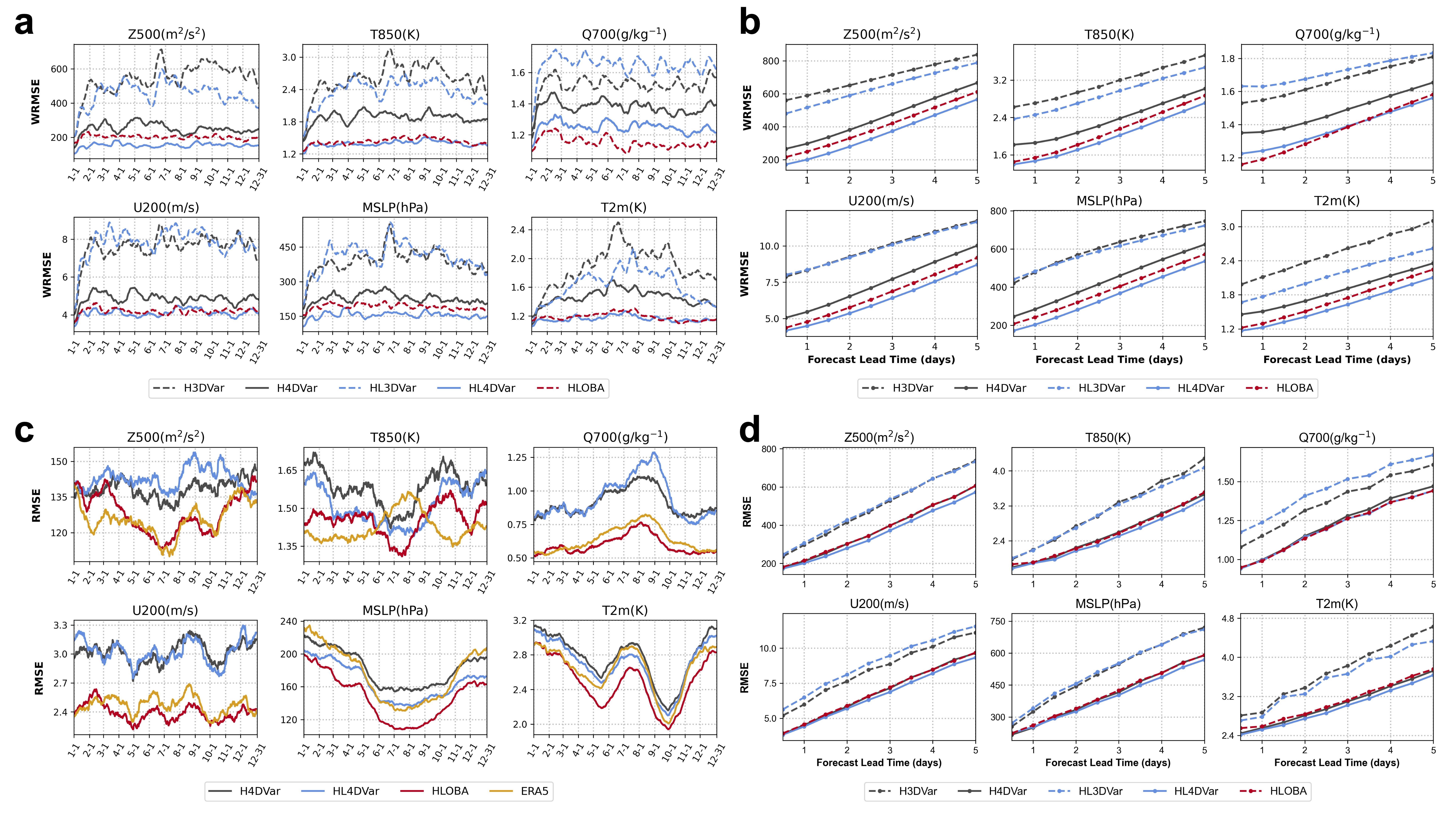}
    \caption{\textbf{Performance comparison of HLOBA and other hybrid DA methods in cycling DA experiments for 2017.} 
    \textbf{a}, Analysis errors from idealized cycling DA experiments, evaluated against ERA5. Observations were sampled from ERA5 using radiosonde and surface observation locations from GDAS at 0000 UTC on 1 January. Four observation times were assimilated in each cycle at 6-hour intervals.
    \textbf{b}, Annual-mean errors for 5-day forecasts initialized from the analyses in a after each assimilation cycle in \textbf{a}, evaluated against ERA5.
    \textbf{c}, Analysis errors from cycling experiments with real observations, evaluated using a withheld 10\% subset of observations not assimilated. Four observation times were assimilated in each cycle at 12-hour intervals. Errors of ERA5 evaluated at the same observation locations are also shown for comparison. For clarity, H3DVar and HL3DVar are not shown.
    \textbf{d}, Annual-mean 5-day forecast errors initialized from the analyses in \textbf{c}, evaluated using all available observations.
    }
    \label{fig:fig2}
\end{figure}

In our idealized experiments, ERA5 is treated as the truth to generate synthetic observations and to evaluate analysis and forecast errors. Synthetic observations are sampled every 6~hours at fixed locations given by the surface and radiosonde stations in GDAS at 00~UTC on 1 January 2017.
Across these experiments (Figure~\ref{fig:fig2}a,b), HLOBA substantially outperforms the other 3D-DA schemes and even surpasses the traditional 4D-DA method, with 15.9\% lower analysis error and 9.2\% lower 5-day forecast error relative to H4DVar.
Its performance is only marginally lower than that of HL4DVar, with HL4DVar yielding 5\% lower analysis and forecast errors.

In experiments with real observations, we assimilate GDAS surface and radiosonde data, withholding 10\% of the measurements for independent verification of the analyses while using the full set to assess forecast performance. As shown in Figure~\ref{fig:fig2}c,d, the overall results align well with the idealized experiments: HLOBA markedly outperforms H3DVar and HL3DVar, and it surpasses traditional 4DVar by reducing the analysis error by 14.9\% and the 5-day forecast error by 0.5\%. Notably, although the forecast error of HLOBA remains 4.2\% higher than that of HL4DVar, the analysis error of HLOBA is 9\% lower.
We also verify ERA5 against the same withheld observations and compare its analysis errors with those from HLOBA. When only surface and radiosonde observations are assimilated, HLOBA outperforms ERA5 for 34 of 69 variables, indicating that HLOBA is more consistent with the assimilated observing system and has the potential to yield analyses that surpass the accuracy of its training data. The main exception is the upper-air temperature, for which ERA5 retains a clear advantage because it benefits from assimilating satellite radiances, which are not included in our experimental design. An expanded version of Figure~\ref{fig:fig2}, covering a more comprehensive set of variables across multiple vertical levels, is provided in Figure~\ref{fig:figS1}-\ref{fig:figS4}.

Beyond analysis and forecast accuracy, computational efficiency is increasingly important for DA as model resolution and observational volume grow. Figure~\ref{fig:fig1}c compares the GPU memory usage, mean wall-clock time per observation time slot assimilated, and the overall performance across different DA schemes in the real-observation experiments on an NVIDIA A100. 
Both 3D-DA and 4D-DA require more than 20~s to assimilate a single observation time slot, with most of the time spent on iterative optimization. Relative to 3D-DA, 4D-DA also requires substantially more GPU memory to store intermediate gradients of the forecast model during optimization, reaching 53.2~GB. In contrast, HLOBA requires only 1.06~s and 10.8~GB of GPU memory. This efficiency arises because HLOBA assimilates observations and background fields in a unified, near-decorrelated latent space, where the analysis can be computed independently for each dimension without iteration.
In summary, HLOBA achieves analysis and forecast skill comparable to that of 4D-DA schemes that enforce model dynamics, while using only a fraction of their computational resources.

\subsection*{Uncertainty quantification for HLOBA analyses}

Quantifying analysis uncertainty in DA with a finite ensemble is inherently challenging as the atmospheric state is extremely high-dimensional. Traditional ensemble DA methods typically requires large ensembles to suppress sampling noise, which is computationally expensive~\cite{houtekamerReviewEnsembleKalman2016a,bannisterReviewOperationalMethods2017a}.
In LDA, however, the near-diagonal background and observation error covariances, $\mathbf{B}_z$ and $\mathbf{R}_z$, allow the latent-space analysis-error variance $\boldsymbol{\sigma}_z^2$ to be diagnosed directly from their diagonal entries. This enables inexpensive, fast uncertainty quantification of the latent analysis without ensemble assimilation.
Building on this property, we estimate the uncertainty of the model-space analysis by perturbing the latent-space analysis along the direction of the latent analysis increment, with an amplitude set by $\boldsymbol{\sigma}_z$, and evaluating the resulting impact in model space. 
This construction can be interpreted as a simple sample from the latent-space uncertainty, and, thanks to the approximately local linearity of the atmospheric AE~\cite{fanPhysicallyConsistentGlobal2026}, the resulting perturbation can be reliably propagated into model space. Further details are provided in the Methods section.

\begin{figure}[ht]
    \centering
    \includegraphics[width=\linewidth]{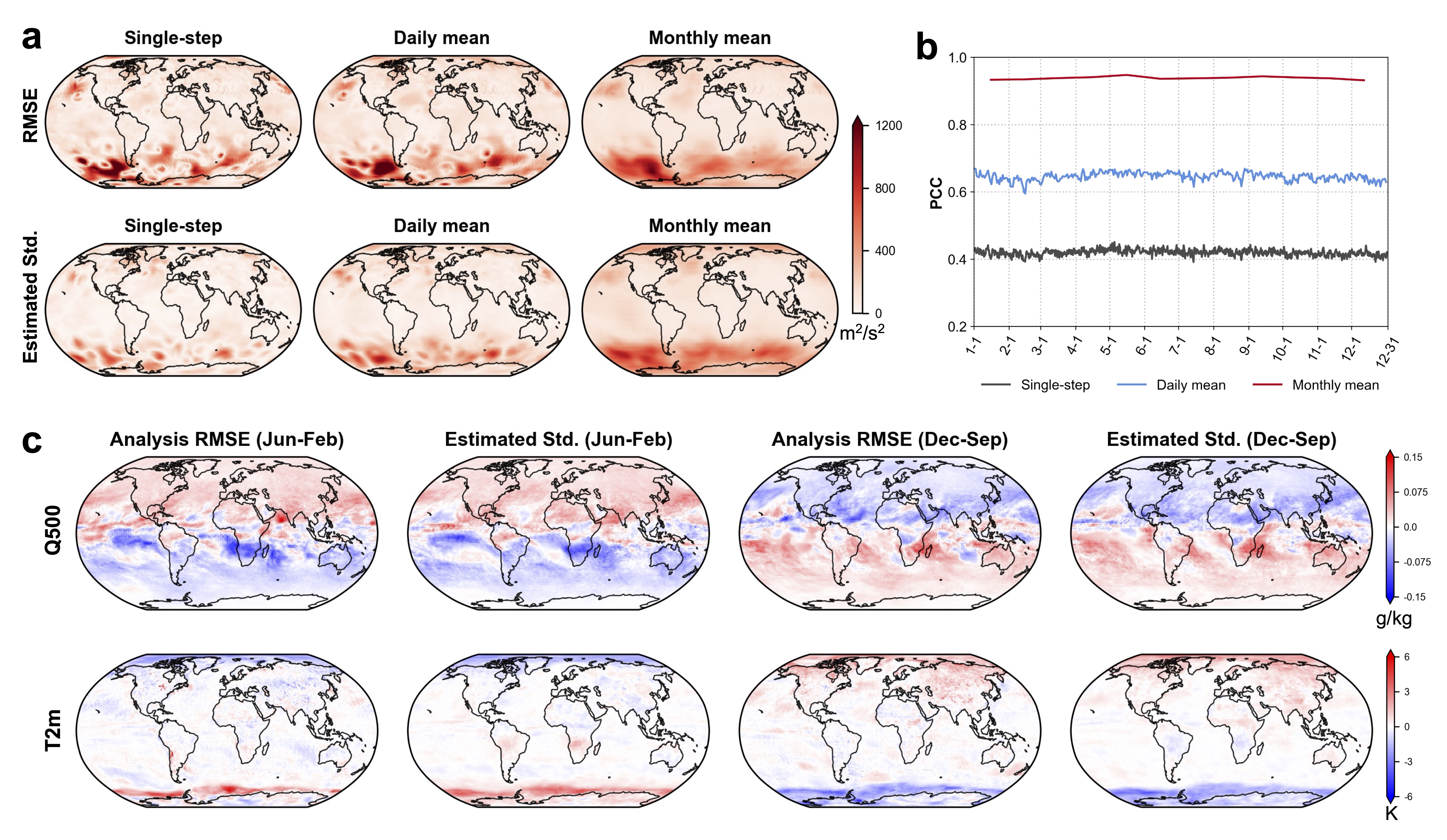}
    \caption{\textbf{Uncertainty estimation of HLOBA analyses in idealized experiments.} 
    \textbf{a}, Examples of analysis RMSE and corresponding estimated standard deviation (Std.) for Z500 at a single time step, daily mean, and monthly mean. Shown are results for 0000 UTC on 1 February 2017, the daily mean on 1 February, and the monthly mean for February 2017.
    \textbf{b}, Pearson correlation coefficients (PCC) between the estimated standard deviation and the true RMSE for single-step, daily-mean, and monthly-mean analyses over 2017.
    \textbf{c}, Seasonal variations in analysis RMSE are reflected in the estimated uncertainty. Results are shown for Q500 and T2m.
    }
    \label{fig:fig3}
\end{figure}

We evaluate the effectiveness of this approach by comparing the diagnosed analysis standard deviation with its root mean square error (RMSE) in an idealized cycling DA experiment. 
For individual DA steps, regions of large error associated with the inhomogeneous observing network are broadly captured, as shown in Figure~\ref{fig:fig3}a, but the spatial patterns are biased, and the annual-mean correlation between the diagnosed uncertainty and the analysis error is about 0.42, as shown in Figure~\ref{fig:fig3}b. However, the agreement improves markedly with temporal averaging, as the correlation increases to around 0.65 for daily means and to roughly 0.94 for monthly means. 
In addition, we find a clear seasonal drift in the monthly-mean analysis error for HLOBA (Figure~\ref{fig:fig3}c), with opposite phases in winter and spring. Despite relying on only three ensemble members, the uncertainty estimates still capture this expected seasonal modulation in both amplitude and spatial structure.

\subsection*{Impact of ensemble information on DA schemes}
To quantify the impact of time-lagged ensembles on each DA scheme, we compare their analysis and forecast errors obtained with and without ensemble information in both the idealized and real-observation experiments. 
We examine three time-lagged ensemble configurations constructed as time-lagged ensembles comprising 3, 6, and 9 members, respectively, and assess each DA method under its individually optimized (best-performing) hyperparameter settings.

\begin{figure}[ht]
    \centering
    \includegraphics[width=\linewidth]{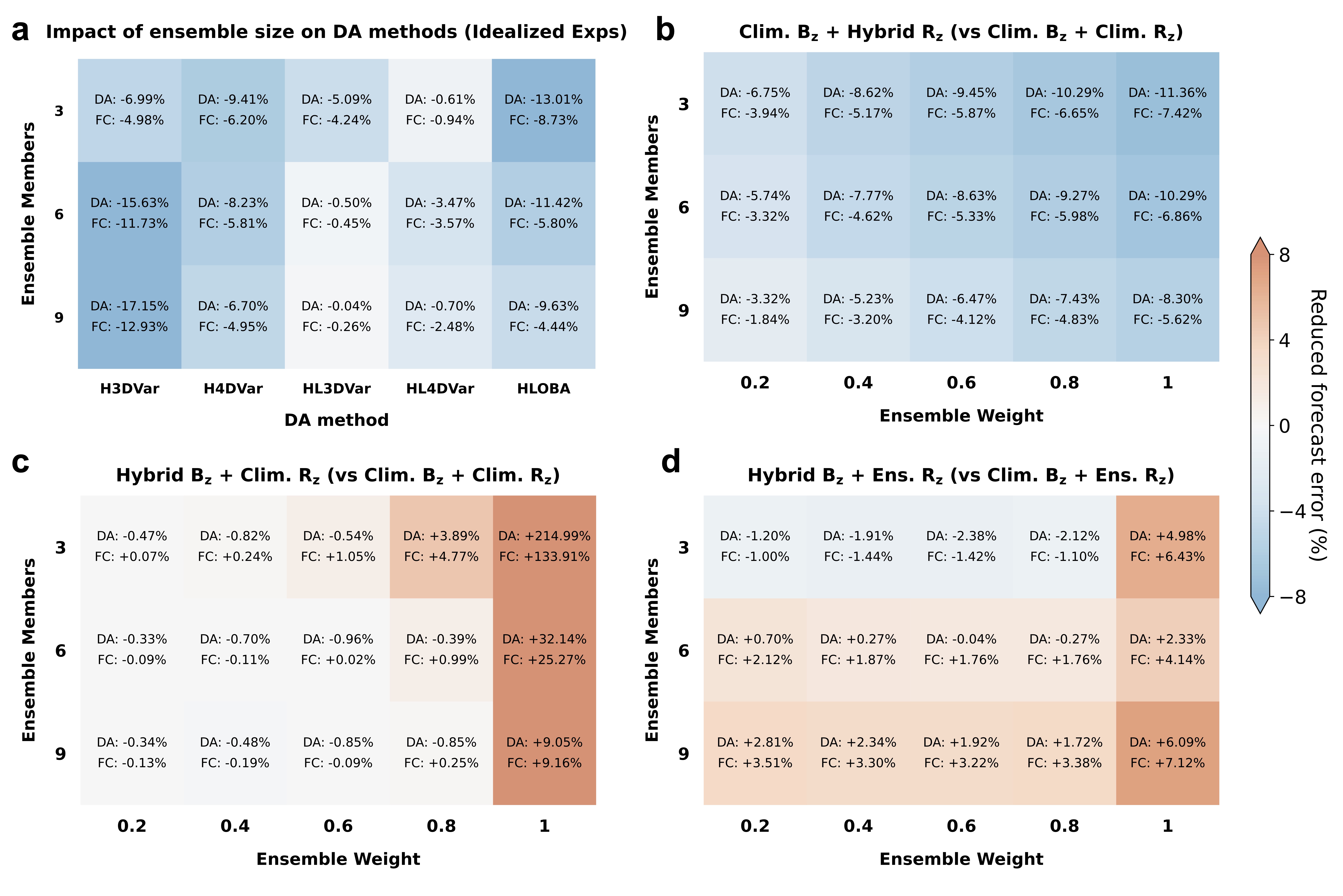}
    \caption{\textbf{Impact of time-lagged ensembles on hybrid DA methods in idealized experiments.} 
    \textbf{a}, Impact of introducing time-lagged ensembles on different hybrid DA methods, relative to their non-ensemble counterparts.
    \textbf{b}, Improvement from using hybrid-ensemble (Hybrid) latent observation error covariance $\mathbf{R}_z$ in HLOBA, relative to a climatological configuration (clim.) without ensemble information.
    \textbf{c}, Same as \textbf{b}, but using hybrid-ensemble latent background error covariance $\mathbf{B}_z$. 
    \textbf{d}, Additional improvement from introducing a hybrid $\mathbf{B}_z$ in HLOBA when $\mathbf{R}_z$ is fully ensemble-based (Ens.).
    Values show mean improvements averaged over the data assimilation (DA) and forecast (FC) stages; colors indicate improvements in the forecast stage only.
    }
    \label{fig:fig4}
\end{figure}

The sensitivity to the ensemble information differs markedly across methods (Figure~\ref{fig:fig4}a). Time-lagged ensembles deliver the largest gains for H3DVar, reducing analysis errors by up to 17.2\% and 12.9\%, respectively.
For H4DVar, the ensembles yield about a 9.4\% reduction in analysis error and a 6.2\% reduction in forecast error. By contrast, their latent-space counterparts improve by $<6\%$ for both the analysis and the forecast.
This indicates that ensemble sampling provides a weaker enhancement of $\mathbf{B}_z$ in latent-space DA than of $\mathbf{B}$ in model-space DA, which is consistent with the conclusions reported in~\cite{melincUnifiedNeuralBackgroundError2026}.
This is because ensembles can introduce highly informative flow-dependent covariances into the climatological $\mathbf{B}$, whereas in intrinsically physically consistent LDA, this information is largely embedded in the latent representation and therefore adds little~\cite{fanNovelLatentSpace2025a,fanPhysicallyConsistentGlobal2026}. 
Nonetheless, HLOBA benefits more from ensemble information than the other LDA schemes.

To explore where this advantage comes from, we perform ablation experiments that isolate the two ensemble-driven components in HLOBA, namely the estimation of the latent-space background and observation error covariances, $\mathbf{B}_z$ and $\mathbf{R}_z$. 
Specifically, we consider configurations in which only one of $\mathbf{B}_z$ or $\mathbf{R}_z$ depends on ensemble statistics, while the other matrix is fixed at its climatological estimate, and then quantify the resulting gains relative to a no-ensemble configuration. 
As shown in Figure~\ref{fig:fig4}b, the ensemble has a strong impact on $\mathbf{R}_z$, as across all tested ensemble sizes the optimal blending weight assigns full weight to the ensemble component, reducing both analysis and forecast errors by up to 11.4\% and 7.4\%, respectively. 
In comparison, for $\mathbf{B}_z$ (Figure~\ref{fig:fig4}c), the benefit is much smaller and only emerges with a well-chosen blending weight, yielding a marginal ($<1\%$) error reduction in both DA and forecast errors.
Accordingly, the optimal strategy for HLOBA is to use a purely ensemble-estimated $\mathbf{R}_z$ together with a hybrid $\mathbf{B}_z$, as validated in Figure~\ref{fig:fig4}d. Interestingly, a hybrid $\mathbf{B}_z$ yields at most a 0.2\% forecast gain when $\mathbf{R}_z$ is climatological, but about a 1.4\% gain when $\mathbf{R}_z$ is ensemble-estimated. This suggests that estimating both $\mathbf{B}_z$ and $\mathbf{R}_z$ from the ensemble provides complementary benefits.

\subsection*{Role of O2Lnet in HLOBA}
\begin{figure}[ht]
    \centering
    \includegraphics[width=\linewidth]{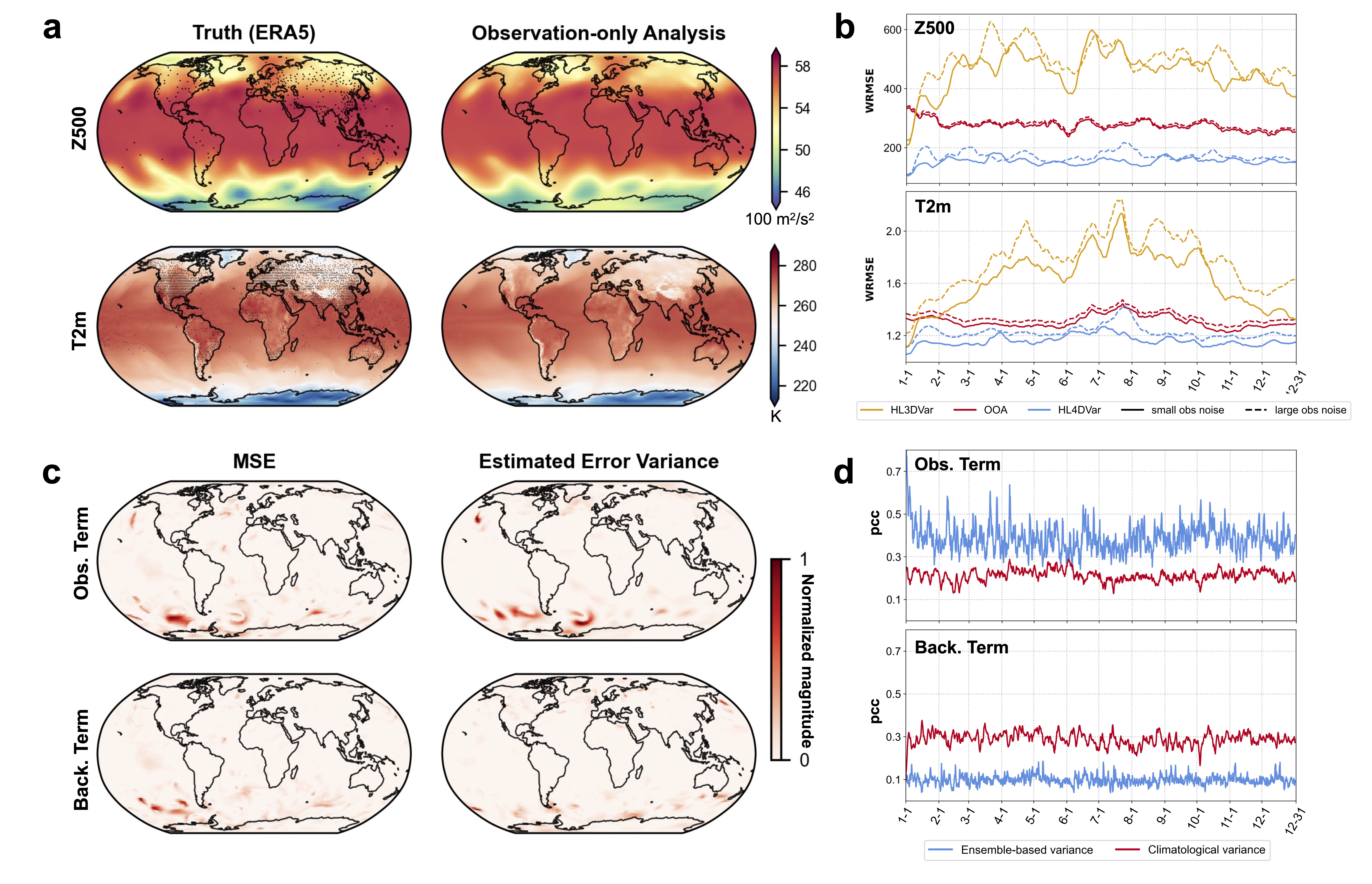}
    \caption{\textbf{Advantages of introducing O2Lnet}, demonstrated under idealized settings.
    \textbf{a}, Illustration of observation-only analysis (OOA) based on O2Lnet, obtained by decoding the latent observation variable $\boldsymbol{z}_o$ produced by O2Lnet. Observations are sampled from ERA5, with observation locations indicated by black dots. Shown are Z500 and T2m at 0000 UTC on 1 February 2017.
    \textbf{b}, Accuracy of OOA and its robustness to observation noise, compared with HL3DVar and HL4DVar. Solid lines show results with observation noise drawn from a zero-mean Gaussian distribution with variance equal to 0.03 times the climatological variance, while dashed lines correspond to a variance of 0.1 times the climatological variance.
    \textbf{c}, Example illustrating the consistency between ensemble-estimated uncertainties of the background and OOA terms and their realized mean squared errors (MSE). Shown is T500 at 0000 UTC on 1 February 2017, with all fields normalized to the range [0, 1].
    \textbf{d}, Pearson correlation coefficients between estimated error variances and realized MSEs for the background and observation terms, $\boldsymbol{z}_b$ and $\boldsymbol{z}_o$. Results based on climatological variance estimates are also shown for comparison.
    }
    \label{fig:fig5}
\end{figure}

By introducing O2Lnet to process observations in an end-to-end manner, HLOBA substantially outperforms L3DVar, even though both are 3D-DA schemes, while also providing robust uncertainty estimates.
To understand how O2Lnet enables these improvements, we examine the latent representation of the observations $\boldsymbol{z}_o$ by decoding them into model space and analyzing the resulting fields $\boldsymbol{x}_o$. In this way, the O2Lnet output $\boldsymbol{x}_o$ can be interpreted as the model-space representation of the observations, which we refer to as the observation-only analysis (OOA).
In the idealized cycling DA experiment, OOA $\boldsymbol{x}_o$ closely matches the ERA5 reanalysis (treated as truth) in both amplitude and spatial structure (Figure~\ref{fig:fig5}a). 
Over the one-year experiment(Figure~\ref{fig:fig5}b), the statistical error of $\boldsymbol{x}_o$ is substantially smaller than that of HL3DVar and approaches the HL4DVar level.
These results indicate that the O2Lnet-derived $\boldsymbol{z}_o$ captures most of the essential features of the atmospheric state, including their spatial and cross-variable correlation, even with sparse observations.
The end-to-end observation treatment is also resilient to noise. When the observation-error standard deviation is increased from 3\% to 10\% of the climatological value, the analysis error of $\boldsymbol{x}_o$ rises by only 8.3\% of the imposed increase, compared with 47.3\% for HL3DVar and 25.6\% for HL4DVar.
Moreover, this end-to-end treatment of observations remains robust to observational noise. When the observation-error standard deviation is increased from 3\% to 10\% of the climatological value, the analysis error of $\boldsymbol{x}_o$ rises by only 8.3\% of the added observation error, compared with 47.3\% for HL3DVar and 25.6\% for HL4DVar.
Additional OOA cases and corresponding statistical results are provided in Figure~\ref{fig:figS5}-\ref{fig:figS7}.

A further advantage of O2Lnet is that it allows uncertainty in the observation-derived latent analysis, $\boldsymbol{z}_o$, to be estimated using background ensembles.
In idealized cycling DA experiments, the ensemble-based estimates of $\boldsymbol{x}_o$ error closely match the realized errors in both magnitude and spatial distribution (Figure~\ref{fig:fig5}c).
To quantify this consistency, Figure~\ref{fig:fig5}d reports the correlation between the ensemble-diagnosed variance of $\boldsymbol{z}_o$ and its realized mean-square error (MSE), and compares it with a climatological estimate. The ensemble-based estimate attains a yearly mean correlation of 0.4, compared with 0.2 for the climatological estimate. This explains why the optimal weight on the climatological component of $\mathbf{R}_z$ is consistently zero in our experiments.

Similarly, we evaluate the ensemble-based errors of the latent background state, $\boldsymbol{z}_b$, against the true $\boldsymbol{z}_b$ errors in the same idealized experiments (Figure~\ref{fig:fig5}c,d). Overall, the time-lagged ensemble produces $\boldsymbol{z}_b$ errors whose spatial distribution differs noticeably from the truth. In the latent space, the correlation between the ensemble-estimated background variance and the realized MSE is only about 0.1, much lower than the 0.3 achieved by the climatological estimate. 
This suggests that the time-lagged ensemble does not capture enough information about the background-error distribution to replace the climatological estimate of $\mathbf{B}_z$, but can still be effective when used to augment the background covariance in a hybrid formulation. These results indicate that the uncertainty estimation capability of HLOBA mainly arises from the introduction of O2Lnet, rather than from the background term.

\section*{Discussion}\label{sec3}

This study presents an efficient and accurate ensemble LDA framework built on two central advances. 
First, we introduce an additional neural network, O2Lnet, that maps observations directly into latent space to fully exploit observational information in an end-to-end manner.
This design enables ensemble-based uncertainty estimation of the latent analysis obtained from observations, which is then combined with the background field in a Bayesian update to produce more accurate analyses and associated uncertainty estimates.
Second, by leveraging the near-diagonal structure of error covariances in the latent space, we use a diagonal approximation for ensemble covariance estimation in LDA, reducing both computational complexity in DA and sampling error in uncertainty quantification.
The results show that, with only three time-lagged ensemble members, our method is competitive with Hybrid L4DVar and even surpasses a GPU-accelerated Hybrid 4DVar, while using only $\sim$3\% of the runtime and $\sim$20\% of the memory.

Beyond computational efficiency, HLOBA offers a key advantage in its flexibility with respect to the forecast model, which is largely absent from existing ML-based DA methods. Although promising, L4DVar and end-to-end analysis methods are, in current practice, restricted to ML forecasting systems: L4DVar explicitly embeds model dynamics in the cost function~\cite{fanPhysicallyConsistentGlobal2026}, whereas end-to-end analysis methods typically require joint tuning with a forecast model to improve accuracy and stability~\cite{chenEndtoendArtificialIntelligence2024,allenEndtoendDatadrivenWeather2025,xiangADAFArtificialIntelligence2025,sunDatatoforecastMachineLearning2025}.
Yet, current ML forecasting systems are concentrated in observation-rich domains with high-quality reanalysis products, most notably global weather prediction, and remain far less mature for regional weather, land-surface and paleoclimate applications, where training data are often less abundant or less reliable. Moreover, L4DVar can be less suitable in domains where model dynamics are less reliable, such as the ocean, as it enforces strong dynamical constraints. These factors are likely to limit the near-term applicability of both L4DVar and end-to-end analysis schemes.
In contrast, as a three-dimensional DA method, HLOBA uses the forecast model only to generate the background state and ensembles, which keeps the forecast model separate from both AE/O2Lnet training and the assimilation algorithm itself. This separation gives HLOBA the potential to be applied across forecasting models and application domains beyond global weather.

Our experiments show that, even when assimilating only surface and radiosonde measurements, HLOBA produces analyses that surpass ERA5 for nearly half of the evaluated variables, while supporting forecast performance comparable to L4DVar. 
Although these analyses may lean too closely toward the observations and thus exhibit some imbalance, they nonetheless suggest that this method has considerable potential. 
When combined with satellite observations, it could outperform ERA5, a capability that current end-to-end schemes cannot offer.
We emphasize that both the training of O2Lnet and the estimation of its practical error statistics rely on simulated observations generated by the observation operator. While this procedure is relatively straightforward for conventional observations, it demands substantial additional engineering effort for indirect observations such as satellite measurements.
In addition, the framework supports multiple plug-and-play O2Lnets, enabling observations from different modalities to be mapped separately into a shared latent space and fused in a Bayesian manner.
Such flexibility is crucial for operational systems, where data gaps caused by transmission failures or instrument outages are commonplace but remain difficult to handle for current end-to-end analysis schemes.

Finally, we note that our use of time-lagged ensembles is a pragmatic compromise necessitated by the current difficulty of generating well-calibrated ensemble spreads from deterministic ML forecast models. Larger and better-calibrated ensembles would refine latent-space uncertainty estimates and thereby further enhance HLOBA's performance. As ML-based probabilistic forecasting systems advance~\cite{kochkovNeuralGeneralCirculation2024,liGenerativeEmulationWeather2024,langAIFSCRPSEnsembleForecasting2024,zhongFuXiENSMachineLearning2025,nathaniel2026generative} and can provide rich ensemble samples at low cost, coupling them with HLOBA to achieve more accurate analyses, robust uncertainty quantification and fully probabilistic forecasts represents a promising avenue for future development.

\section*{Methods}\label{sec4}
\subsection*{Training dataset and observations}
We use the ERA5 reanalysis dataset~\cite{carrassiDataAssimilationGeosciences2018} to train the AE, O2Lnet, and the forecast model. We regrid ERA5 to a 1.40625° global grid and use four surface and five upper-air variables on 13 pressure levels (50, 100, 150, 200, 250, 300, 400, 500, 600, 700, 850, 925, and 1000 hPa), yielding a 69 × 256 × 128 atmospheric model state. Specifically, the surface variables are 2-meter air temperature(T2m), 10-meter winds (U10, V10), and mean sea-level pressure (MSL); the upper-air variables are geopotential height (Z), temperature (T), zonal wind (U), meridional wind (V), and specific humidity (Q). ERA5 reanalysis data from 1979 to 2015 are used to train the models, with 2016 reserved for validation and 2017 for testing and DA experiments.

To provide an initial proof of concept for HLOBA, we only consider a conventional-observation configuration, assimilating surface and radiosonde data from the 2017 GDAS archive (BUFR types ``ADPSFC'' and ``ADPUPA''). The observations are interpolated to the model state grid for simplicity. Surface stations at high elevation are treated as upper-air observations by vertically interpolating them to the corresponding pressure level. When multiple reports fall into the same grid cell, they are averaged to reduce sampling noise. In the real-observation assimilation experiments, a simple quality-control step is applied: observations whose departures from the ERA5 analysis exceed a threshold are discarded to maintain DA stability. For each of the 69 variables, the threshold is defined as the mean absolute difference between ERA5 analyses 48 hours apart, computed over 2016. The thresholds for surface variables are inflated by a factor of four, given that the representativeness and interpolation errors associated with terrain effects are much larger than for upper-air variables.

\subsection*{Forecast Model}
We train the ML-based forecast model based on the architecture of FengWu~\cite{chenOperationalMediumrangeDeterministic2025}, an advanced medium-range forecast model with predictability extending beyond 10 days. However, as in many deterministic forecast models, long-lead forecasts from the original FengWu become increasingly blurred, which is physically unrealistic \cite{nathaniel2024chaosbench,fanIncorporatingMultivariateConsistency2025}. Such degraded dynamics may impair 4D-DA methods (4DVar and L4DVar) that rely on the forecast model as a dynamical constraint. 
To alleviate this, we follow the strategy proposed in~\cite{fanIncorporatingMultivariateConsistency2025}, in which the model-space loss is replaced by a latent-space loss during rollout tuning with up to 12 steps. This encourages the model to maintain more realistic spatial structure while improving its long-range performance. In addition, to support 4D-DA, we train the model to take a single atmospheric state as input, rather than the two consecutive states used by many existing approaches~\cite{lamLearningSkillfulMediumrange2023,chenFuXiCascadeMachine2023,kochkovNeuralGeneralCirculation2024,priceProbabilisticWeatherForecasting2025,chenOperationalMediumrangeDeterministic2025}.
Further details of the model architecture and latent-space-constrained training strategy are given in ~\cite{chenOperationalMediumrangeDeterministic2025,fanIncorporatingMultivariateConsistency2025}.

\subsection*{AE and O2Lnet}
The AE defines the atmospheric latent space that underpins LDA. It comprises an encoder $E(\cdot)$ and a decoder $D(\cdot)$, which map model-space states $\boldsymbol{x}$ to latent representations $\boldsymbol{z}$ and back, respectively. To encourage $\boldsymbol{z}$ to preserve the essential information in $\boldsymbol{x}$, we train the AE by minimizing a mean-squared reconstruction loss, $\left\lVert \boldsymbol{x} - D\!\left(E(\boldsymbol{x})\right) \right\rVert_2^2$.
The AE architecture follows our previous work~\cite{fanPhysicallyConsistentGlobal2026}. The encoder begins with a $4 \times 4$ patch embedding that partitions the input field into $64 \times 32$ tokens, each represented by a 1024-dimensional vector augmented with learnable absolute positional encodings (APE)~\cite{dosovitskiyImageWorth16x162021}.
The resulting token sequence is processed by 12 Swin Transformer blocks~\cite{liuSwinTransformerHierarchical2021} with 16 attention heads. A shared fully connected network (FCN) is then applied token-wise to project the features into a latent tensor of shape $69 \times 64 \times 32$. The decoder mirrors the encoder architecture, with a final FCN applied to each latent token and reshaped to recover the original model-space structure.

In HLOBA, O2Lnet is designed to map observations directly into the AE latent space and is therefore trained after the AE is pretrained and frozen.
We trained O2Lnet using simulated observations generated by the observation operator $\mathcal{H}$. 
Specifically, for any model-space training sample $\boldsymbol{x}$, the simulated observations $\mathcal{H}(\boldsymbol{x})$ are used as the input and the corresponding latent representation ${E}(\boldsymbol{x})$ as the target. To account for time-varying observation locations and errors, we provide O2Lnet with an accompanying quality-mask tensor $\boldsymbol{m}_{q}$ that matches the model-state dimensions and encodes both observation coverage and reliability as an additional input.
The mask values are assigned heuristically based on time and spatial mismatches, whether interpolation is applied, and GDAS quality-control flags, with values closer to 1 indicating higher-quality observations and 0 denoting missing or unusable data. 
During training, we use the quality mask to modulate the noise added to the simulated observations, injecting larger perturbations for lower-quality values so that O2Lnet can learn to account for observation errors.

To account for regional differences in mapping difficulty that result from uneven observational coverage, we weight the training loss with a spatial weighting map $\boldsymbol{w}(\boldsymbol{m}_{q})$ during O2Lnet training. Specifically, we smooth the quality-mask tensor $\boldsymbol{m}_{q}$ with a horizontal Gaussian kernel, average the result across variables, reshape it to the spatial resolution of the latent space, and then rescale it to $[0.5, 1]$ to to model the influence of observation quality and spatial distribution.
Therefore, the O2Lnet is trained by minimizing 
\begin{equation}
\label{eq:O2L_loss}
\mathcal{L}_{\mathrm{O2L}}
= \left\lVert \boldsymbol{w}(\boldsymbol{m}_{q}) \odot
[( \mathrm{O2L}\!\left(\mathcal{H}(\boldsymbol{x}) + \boldsymbol{\epsilon}(\boldsymbol{m}_{q}),\, \boldsymbol{m}_{q}\right)
- E(\boldsymbol{x})]) \right\rVert_2^2,
\end{equation}
where $\mathrm{O2L}(\cdot,\cdot)$ denotes O2Lnet and $\odot$ denotes element-wise multiplication.
$\mathcal{H}(\boldsymbol{x}) + \boldsymbol{\epsilon}(\boldsymbol{m}_{q})$ denotes the simulated observations corrupted with mask-dependen perturbations $\boldsymbol{\epsilon}(\boldsymbol{m}_{q})$. We implement O2Lnet with a gated convolutional neural network~\cite{yuFreeFormImageInpainting2019} that fuses the observations and the quality mask to produce $64\times32$ tokens, each with a feature dimension of 1024, followed by a Swin Transformer and an FCN head that maps the output to the latent-space resolution of $69\times64\times32$.

Both the AE and O2Lnet were trained for 30 epochs using the AdamW optimizer~\cite{loshchilovDecoupledWeightDecay2019}. We used a learning rate of $2\times10^{-4}$ with linear warm-up followed by cosine decay. Notably, both models are trained in an unsupervised/self-supervised manner using only model-space data and do not rely on any real observations.

\subsection*{HLOBA method}
With the encoder and O2Lnet, both the background field and the observations can be mapped into the same latent space, denoted by $\boldsymbol{z}_b$ and $\boldsymbol{z}_o$, respectively. Under the Gaussian error assumption, with latent background- and observation-error covariances $\mathbf{B}_z$ and $\mathbf{R}_z$, the Best Linear Unbiased Estimator (BLUE) for the latent analysis $\boldsymbol{z}_a$ is
\begin{align}
&\boldsymbol{z}_a = \boldsymbol{z}_b + \mathbf{K}_z(\boldsymbol{z}_o-\boldsymbol{z}_b), \label{eq:BLUE_latent}\\
&\mathbf{K}_z     = \mathbf{B}_z(\mathbf{B}_z+\mathbf{R}_z)^{-1}, \label{eq:Kz}\\
&\mathbf{A}_z     = (\mathbf{I}-\mathbf{K}_z)\mathbf{B}_z, \label{eq:Az}
\end{align}
where $\mathbf{K}_z$ denotes the latent-space Kalman gain and $\mathbf{A}_z$ is the estimated error covariance of the latent analysis $\boldsymbol{z}_a$. The final analysis in model space is obtained by decoding the latent analysis, i.e., $\boldsymbol{x}_a = D(\boldsymbol{z}_a)$. 
In practice, since $\mathbf{B}_z$ and $\mathbf{R}_z$ are nearly diagonal, we use the following approximate formulas to compute the latent analysis and its estimated variance for each dimension $i$:
\begin{gather}
z_{a,i} = z_{b,i} + \frac{(\mathbf{B}_z)_{ii}}{(\mathbf{B}_z)_{ii}+(\mathbf{R}_z)_{ii}}\big(z_{o,i}-z_{b,i}\big), \label{eq:za_i}\\[6pt]
(\mathbf{A}_z)_{ii} = \frac{(\mathbf{B}_z)_{ii}(\mathbf{R}_z)_{ii}}{(\mathbf{B}_z)_{ii}+(\mathbf{R}_z)_{ii}}. \label{eq:A_i}
\end{gather}

When the AE has negligible reconstruction error and the decoder is approximately affine, the model-space analysis error $\boldsymbol{e}_{x_a}$ can be approximated from the latent-space error $\boldsymbol{e}_{z_a}$ as:
\begin{equation}
\label{eq:ex_from_ez}
\boldsymbol{e}_{x_a} \approx D\!\left(\boldsymbol{z}_a+\boldsymbol{e}_{z_a}\right)-D\!\left(\boldsymbol{z}_a\right).
\end{equation}
As these assumptions are well satisfied by the AE for global atmosphere~\cite{fanPhysicallyConsistentGlobal2026}, we approximate the latent-space analysis error $\boldsymbol{e}_{z_a}$ by combining the magnitude of the estimated latent standard deviation, $\boldsymbol{\sigma}_{z_a}=\sqrt{\mathrm{diag}(\mathbf{A}_z)}$, with the direction of the HLOBA assimilation increment, $\mathrm{sign}(\boldsymbol{z}_a-\boldsymbol{z}_b)$. We then estimate the model-space analysis variance (the diagonal of $\mathbf{A}_x=\mathbb{E}[\boldsymbol{e}_{x_a}\boldsymbol{e}_{x_a}^\top]$) as:
\begin{equation}
\label{eq:Ax_diag_sign}
\mathrm{diag}(\mathbf{A}_x)\approx
\left(
\frac{1}{2}({D\!\left(\boldsymbol{z}_a+\boldsymbol{\sigma}_{z_a}\odot \mathrm{sign}(\boldsymbol{z}_a-\boldsymbol{z}_b)\right)
-
D\!\left(\boldsymbol{z}_a-\boldsymbol{\sigma}_{z_a}\odot \mathrm{sign}(\boldsymbol{z}_a-\boldsymbol{z}_b)\right)}
\right)^{\odot 2},
\end{equation}
where $\odot$ denotes element-wise multiplication and $(\cdot)^{\odot 2}$ denotes element-wise squaring.

\subsection*{Estimation of Error covariance matrix}
Accurate estimation of $\mathbf{B}_z$ and $\mathbf{R}_z$ is essential for HLOBA. Following traditional DA, we consider two latent background-error covariance models: a climatological $\mathbf{B}_z^{\mathrm{clim}}$ derived using the NMC method~\cite{parrishNationalMeteorologicalCenters1992} and a flow-dependent $\mathbf{B}_z^{\mathrm{ens}}$ estimated from ensemble samples. The $\mathbf{B}_z^{\mathrm{clim}}$ is defined as:
\begin{equation}
\textbf{B}_z^{clim}
\approx  \frac{1}{2} \left \langle  [E(\boldsymbol{x}^{48})-E(\boldsymbol{x}^{24})][E(\boldsymbol{x}^{48})-E(\boldsymbol{x}^{24})]^\mathrm{T}  \right \rangle, \\
\end{equation}  
where $\langle \cdot \rangle$ denotes an average over samples, $\boldsymbol{x}^{24}$ and $\boldsymbol{x}^{48}$ are 24-h and 48-h model forecasts valid at the same verification time. The $\mathbf{B}_z^{\mathrm{ens}}$ is formulated as:
\begin{equation}
\mathbf{B}_z^{\mathrm{ens}}
\approx
\frac{1}{N_e-1}\sum_{n=1}^{N_e}
\left(E(\boldsymbol{x}^{(n)})-\overline{E(\boldsymbol{x})}\right)
\left(E(\boldsymbol{x}^{(n)})-\overline{E(\boldsymbol{x})}\right)^{\mathrm{T}},
\end{equation}  
where $N_e$ is the ensemble size, $n$ indexes the ensemble members, and $\overline{E(\boldsymbol{x})}=\frac{1}{N_e}\sum_{m=1}^{N_e}E(\boldsymbol{x}^{(m)})$ denotes the ensemble mean in latent space. 
HLOBA adopts a hybrid strategy, representing $\mathbf{B}_z$ as a linear combination of the climatological and ensemble estimates as:
\begin{equation}
\label{eq:Bz_hybrid}
\mathbf{B}_z
=
\alpha_{\mathrm{ens}}\,\mathbf{B}_z^{\mathrm{ens}}
+\left(1-\alpha_{\mathrm{ens}}\right)\mathbf{B}_z^{\mathrm{clim}},
\end{equation}
where $\alpha_{\mathrm{ens}}\in[0,1]$ controls the weight assigned to the ensemble estimate background error. 

Similarly, we estimate a climatological $\mathbf{R}_z^{clim}$ average over a large archive of historical samples $\boldsymbol{x}$ as:
\begin{equation}
\textbf{R}_z^{clim}
\approx  \frac{1}{2} \left \langle [\mathrm{O2L}(\mathcal{H}(\boldsymbol{x}) + \boldsymbol{\epsilon}(\boldsymbol{m}_{q})) -E(x)][\mathrm{O2L}(\mathcal{H}(\boldsymbol{x}) + \boldsymbol{\epsilon}(\boldsymbol{m}_{q})) -E(x)]^\mathrm{T}  \right \rangle,
\end{equation} 
and a flow-dependent ensemble estimate $\mathbf{R}_z^{ens}$ as:
\begin{equation}
\label{eq:Rz_ens}
\mathbf{R}_z^{\mathrm{ens}}
\approx
\frac{1}{N_e-1}\sum_{n=1}^{N_e}
\left(\mathrm{O2L}(\mathcal{H}(\boldsymbol{x}^{(n)}) + \boldsymbol{\epsilon}(\boldsymbol{m}_{q})) - E(\boldsymbol{x}^{(n)}))\right)
\left(\mathcal{H}(\boldsymbol{x}^{(n)}) + \boldsymbol{\epsilon}(\boldsymbol{m}_{q})) - E(\boldsymbol{x}^{(n)})\right)^{\mathrm{T}}.
\end{equation}
The two are then combined in a hybrid form to specify $\mathbf{R}_z$
\begin{equation}
\label{eq:Rz_hybrid}
\mathbf{R}_z
=
\beta_{\mathrm{ens}}\,\mathbf{R}_z^{\mathrm{ens}}
+\left(1-\beta_{\mathrm{ens}}\right)\mathbf{R}_z^{\mathrm{clim}},
\end{equation}
with $\beta_{\mathrm{ens}}\in[0,1]$ controlling the relative weight of the ensemble contribution.

Benefiting from the decorrelation of errors in latent space, the covariance calculations can be simplified by retaining only the diagonal components in the implementation. In addition, these covariance matrices can be further tuned in practice using multiplicative inflation factors to mitigate estimation errors. 

In our implementation, $\boldsymbol{x}^{24}$ and $\boldsymbol{x}^{48}$ are computed at 6 hour intervals throughout 2016, with 6 hourly ERA5 reanalysis fields used as $\boldsymbol{x}$. The resulting climatological estimates of $\mathbf{B}_z$ and $\mathbf{R}_z$ are held fixed for all experiments in 2017. Ensemble members $\boldsymbol{x}^{(n)}$ are constructed with a time-lagged ensemble approach. Hyperparameters governing ensemble inflation and ensemble weighting are tuned using forecast-based evaluation of experiments in November and December 2016 and then held fixed for all experiments in 2017.

\subsection*{Other DA methods}
We compare HLOBA with variational DA methods in both model space and latent space, which solve for the analysis by minimizing a variational cost function. The cost function of the three-dimensional variational method in model space (3DVar)~\cite{courtierECMWFImplementationThreedimensional1998} is formulated as:
\begin{equation}
\label{eq:3dvar_cost}
J(\boldsymbol{x})
=\frac{1}{2}\left(\boldsymbol{x}-\boldsymbol{x}_b\right)^{\top}\mathbf{B}^{-1}\left(\boldsymbol{x}-\boldsymbol{x}_b\right)
+\frac{1}{2}\left(\mathcal{H}(\boldsymbol{x})-\boldsymbol{y}\right)^{\top}\mathbf{R}^{-1}\left(\mathcal{H}(\boldsymbol{x})-\boldsymbol{y}\right),
\end{equation}
where $\textbf{B}$ and $\textbf{R}$ represent the error covariance matrix for the background field $\boldsymbol{x_b}$ and observations $\boldsymbol{y}$, respectively.
The cost function for the four-dimensional variational method (4DVar)~\cite{courtierStrategyOperationalImplementation1994}, which explicitly accounts for model dynamics, is formulated as:
\begin{equation}
\label{eq:4dvar_cost}
J(\boldsymbol{x})=\frac{1}{2}(\boldsymbol{x}-\boldsymbol{x_b})^\mathrm{T} \textbf{B}^{-1}(\boldsymbol{x}-\boldsymbol{x_b}) + \frac{1}{2} \sum_{i=0}^{n} (\boldsymbol{y_i}-\mathcal{H}(M_{0\rightarrow i}(\boldsymbol{x})))^\mathrm{T} \textbf{R}_i^{-1} (\boldsymbol{y_i}-\mathcal{H}(M_{0\rightarrow i}(\boldsymbol{x}))),
\end{equation}
where $i=0,1,\ldots,n$ indexes the assimilation times, $\mathcal{M}_{0\rightarrow i}$ denotes the model forecast operator from the initial time to time $i$, and $\boldsymbol{x}_i=\mathcal{M}_{0\rightarrow i}(\boldsymbol{x}_0)$ is the model trajectory propagated from the initial condition $\boldsymbol{x}_0$.

The cost function for the latent-space 3DVar (L3DVar)~\cite{melinc3DVarDataAssimilation2024} is formulated as:
\begin{equation}
J(\boldsymbol{z})=\frac{1}{2}(\boldsymbol{z}-\boldsymbol{z_b})^\mathrm{T} \textbf{B}_z ^{-1}(\boldsymbol{z}-\boldsymbol{z_b}) + \frac{1}{2} (\boldsymbol{y}-\mathcal{H}(D(\boldsymbol{z})))^\mathrm{T} \textbf{R}^{-1} (\boldsymbol{y}-\mathcal{H}(D(\boldsymbol{z}))),
\end{equation}
and the cost function for the latent-space 4DVar (L4DVar)~\cite{fanPhysicallyConsistentGlobal2026} can be expressed as: 
\begin{equation}
J(\boldsymbol{z})=\frac{1}{2}(\boldsymbol{z}-\boldsymbol{z_b})^\mathrm{T} \textbf{B}_z^{-1}(\boldsymbol{z}-\boldsymbol{z_b}) + \frac{1}{2} \sum_{i=0}^{n} (\boldsymbol{y_i}-\mathcal{H}(M_{0\rightarrow i}(D(\boldsymbol{z})))^\mathrm{T} \textbf{R}_i^{-1} (\boldsymbol{y_i}- \mathcal{H}(M_{0\rightarrow i}(D(\boldsymbol{z})))).
\end{equation}

In practice, we assume that observation errors are uncorrelated and assign each variable a standard deviation equal to 0.03 times its annual standard deviation, yielding a diagonal $\mathbf{R}$ matrix. For the background error covariance matrix, we consider both a climatological estimate and a flow-dependent ensemble estimate and therefore adopt a hybrid formulation that combines these two components. For L3DVar and L4DVar, the latent space background covariance $\mathbf{B}_z$ is identical to that used in HLOBA. For 3DVar and 4DVar, both the climatological and flow-dependent $\mathbf{B}$ are simplified using a control variable formulation, following the NCAR-developed GE\_BE\_2.0 method~\cite{descombesGeneralizedBackgroundError2015}. Hyperparameters for these methods were tuned using experiments in November and December 2016 and then held fixed for all experiments in 2017. All cost functions were minimized automatically in PyTorch using gradient-based optimization with backpropagation. We solve 3DVar with L-BFGS, as its objective is strictly convex. By contrast, the remaining cost functions include neural networks and lead to non-convex objectives, requiring a stochastic optimizer like Adam~\cite{kingmaAdamMethodStochastic2017}.

\subsection*{Metrics}
To evaluate analysis and forecast errors, we report the latitude-weighted root mean square error (WRMSE) for idealized experiments and the standard root mean square error (RMSE) for real-observation experiments. The WRMSE for variable $c$ of the model field $\boldsymbol{x}$ is defined as
\begin{equation}
\label{eq:wrmse}
\operatorname{WRMSE}(\boldsymbol{x},\boldsymbol{x}_{\mathrm{truth}},c)
=
\sqrt{
\frac{1}{HW}
\sum_{h,w}\
\frac{\cos(\alpha_{h,w})}{\sum_{h^{\prime}=1}^{H}\cos(\alpha_{h^{\prime},w})}
\left(\boldsymbol{x}^{c,h,w}-\boldsymbol{x}_{\mathrm{truth}}^{c,h,w}\right)^{2}},
\end{equation}
where $c$, $h$, and $w$ index the variable, latitude, and longitude, respectively; $\alpha_{h,w}$ is the latitude of of point $(h,w)$; and $H$ and $W$ denote the numbers of grid points in the latitudinal and longitudinal directions. The RMSE for variable $c$ in real-observation experiments is defined as
\begin{equation}
\label{eq:rmse_obs}
\operatorname{RMSE}(\boldsymbol{x},\boldsymbol{y},c)
=
\sqrt{\frac{1}{N}\sum_{i=1}^{N}\left(\mathcal{H}(\boldsymbol{x})^{c,i}-\boldsymbol{y}^{c,i}\right)^2},
\end{equation}
where $i$ indexes the observations of variable $c$, $y^{c,i}$ denotes the $i$-th observation, and $N$ is the total number of observations for that variable.

To quantify the accuracy of the estimated analysis uncertainty in the idealized experiments, we compute the Pearson correlation between the analysis RMSE vector $\boldsymbol{e}_{x_a}$ and the estimated analysis standard deviation $\widehat{\boldsymbol{\sigma}}_{x_a}$, evaluated at each analysis step as well as over daily and monthly windows:
\begin{equation}
\label{eq:uncert_corr}
\rho_{\boldsymbol{x}}
=
\mathrm{corr}\!\left(
\boldsymbol{e}_{x_a},
\widehat{\boldsymbol{\sigma}}_{x_a}
\right).
\end{equation}
In addition, to evaluate latent-space uncertainty estimates derived from either ensembles or climatology, we compute the Pearson correlation between the realized element-wise squared error of $\boldsymbol{z}_b$ (or $\boldsymbol{z}_o$) and the corresponding estimated latent variance as
\begin{align}
\label{eq:latent_uncert_corr}
\rho_{z_b}^{clim}
&=
\mathrm{corr}\!\left(
\left(\boldsymbol{z}_b-\boldsymbol{z}_{\mathrm{truth}}\right)^{\odot 2},
\ \mathrm{diag}\!\left({\mathbf{B}}_z^{clim}\right)
\right), \\
\rho_{z_b}^{ens}
&=
\mathrm{corr}\!\left(
\left(\boldsymbol{z}_b-\boldsymbol{z}_{\mathrm{truth}}\right)^{\odot 2},
\ \mathrm{diag}\!\left({\mathbf{B}}_z^{ens}\right)
\right), \\
\rho_{z_o}^{clim}
&=
\mathrm{corr}\!\left(
\left(\boldsymbol{z}_o-\boldsymbol{z}_{\mathrm{truth}}\right)^{\odot 2},
\ \mathrm{diag}\!\left({\mathbf{R}}_z^{clim}\right)
\right),\\
\rho_{z_o}^{ens}
&=
\mathrm{corr}\!\left(
\left(\boldsymbol{z}_o-\boldsymbol{z}_{\mathrm{truth}}\right)^{\odot 2},
\ \mathrm{diag}\!\left({\mathbf{R}}_z^{ens}\right)
\right).
\end{align}

\section*{Declarations}

\textbf{Acknowledgements}.
H. F., J. N., and P. G. acknowledge support from the National Science Foundation (NSF) Science and Technology Center (STC) Learning the Earth with Artificial Intelligence and Physics (LEAP, Award \#2019625).

\textbf{Competing interests}
The authors declare no competing interests.

\textbf{Code availability}
The neural network model is developed using PyTorch. Codes and model checkpoints used in this study will be available at \url{https://github.com/hangfan99/HLOBA} at time of publication.

\bibliographystyle{unsrt}
\bibliography{references}  

\clearpage
\newpage

\section*{Supplementary materials}
\setcounter{figure}{0}
\renewcommand{\thefigure}{S\arabic{figure}}

\begin{figure}[h]
    \centering
    \includegraphics[width=\linewidth]{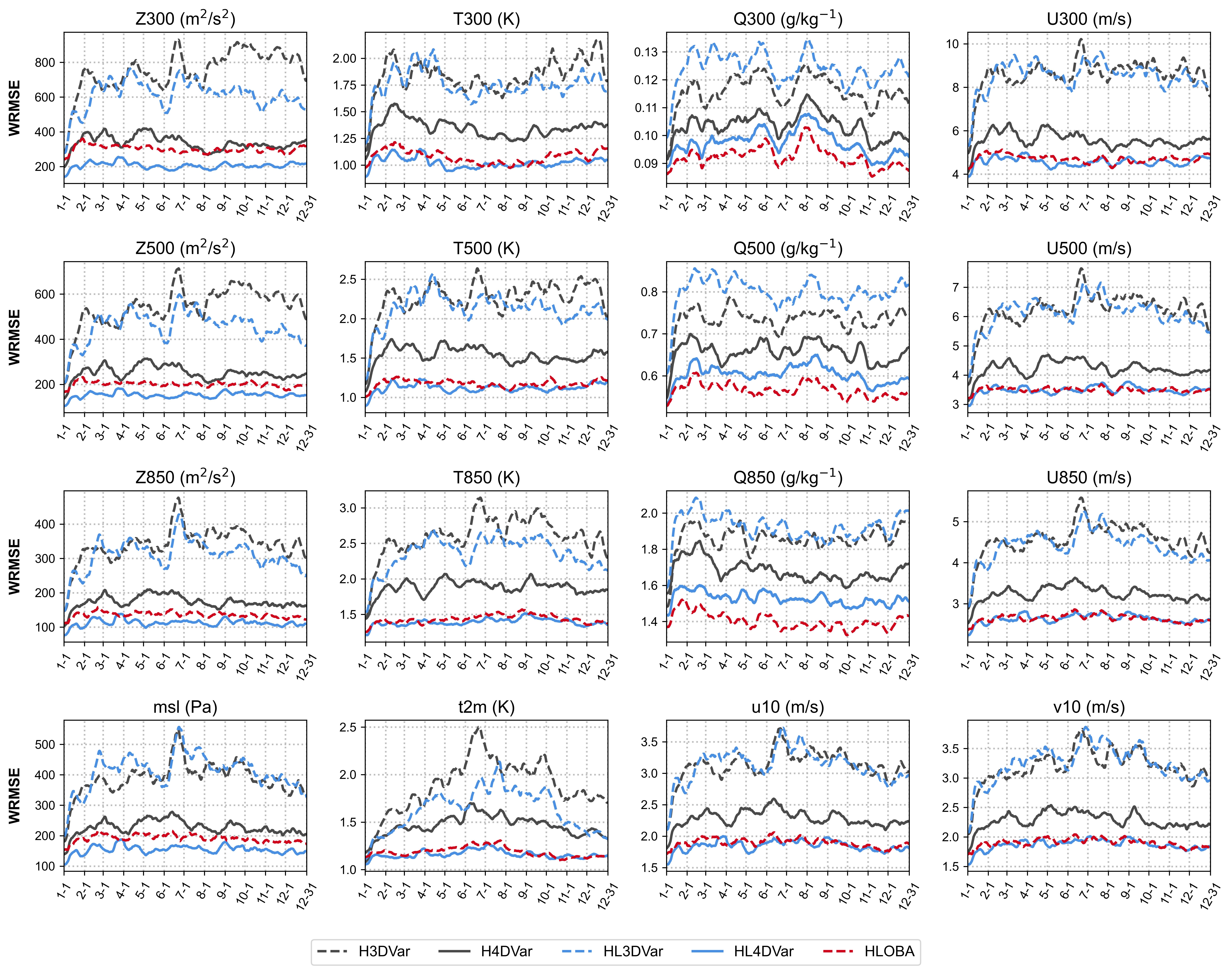}
    \caption{\textbf{Extended analysis performance comparison of HLOBA and other hybrid DA methods in idealized cycling experiments for 2017.} This figure extends Fig.~\ref{fig:fig2}a by reporting WRMSE for variables at 300\,hPa, 500\,hPa, 850\,hPa, and the surface, evaluated against ERA5.}
    \label{fig:figS1}
\end{figure}

\begin{figure}[p]
    \centering
    \includegraphics[width=\linewidth]{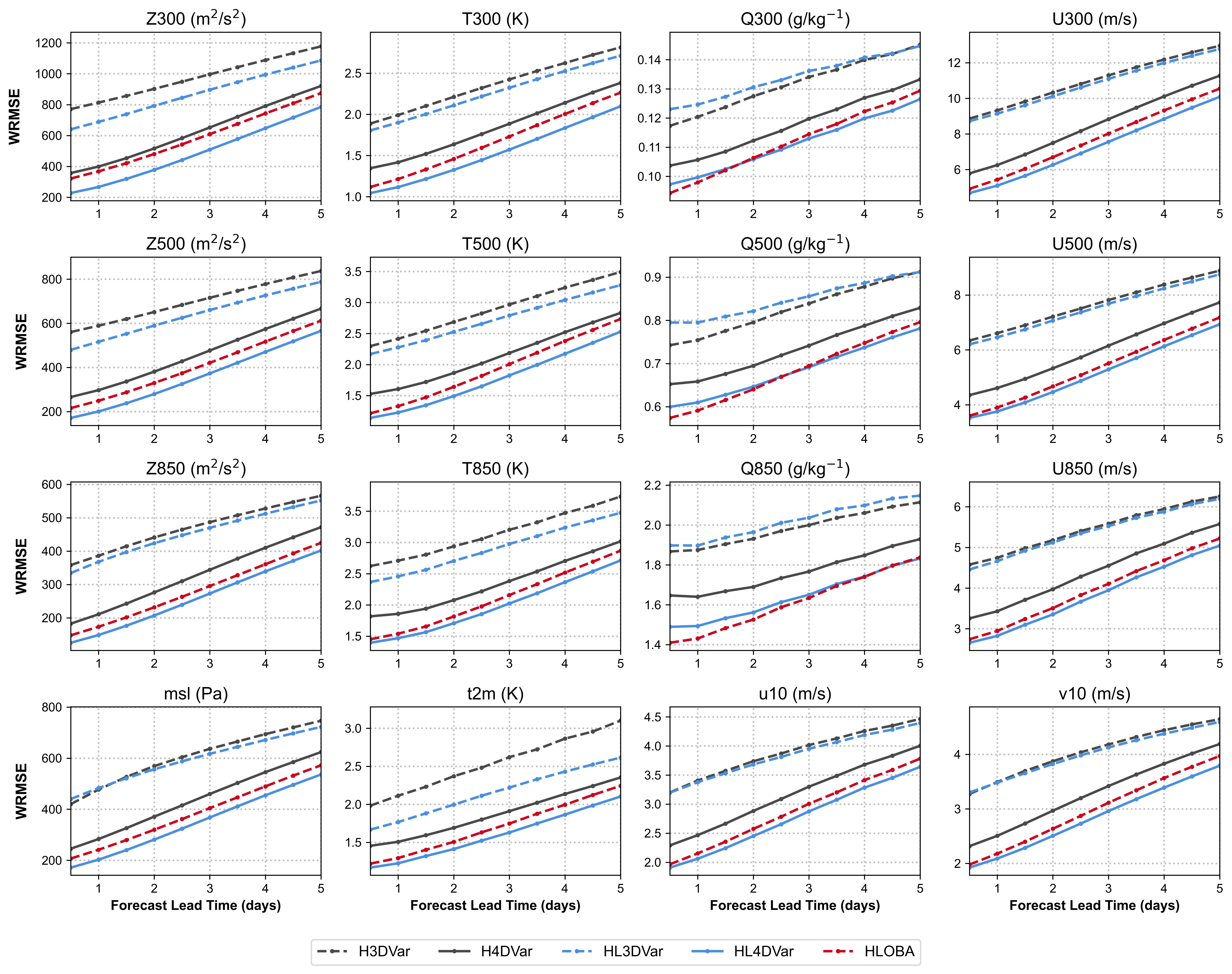}
    \caption{\textbf{Extended forecast performance comparison of HLOBA and other hybrid DA methods in idealized cycling experiments for 2017.} This figure extends Fig.~\ref{fig:fig2}b by reporting WRMSE for variables at 300\,hPa, 500\,hPa, 850\,hPa, and the surface, evaluated against ERA5.}
    \label{fig:figS2}
\end{figure}

\begin{figure}[p]
    \centering
    \includegraphics[width=\linewidth]{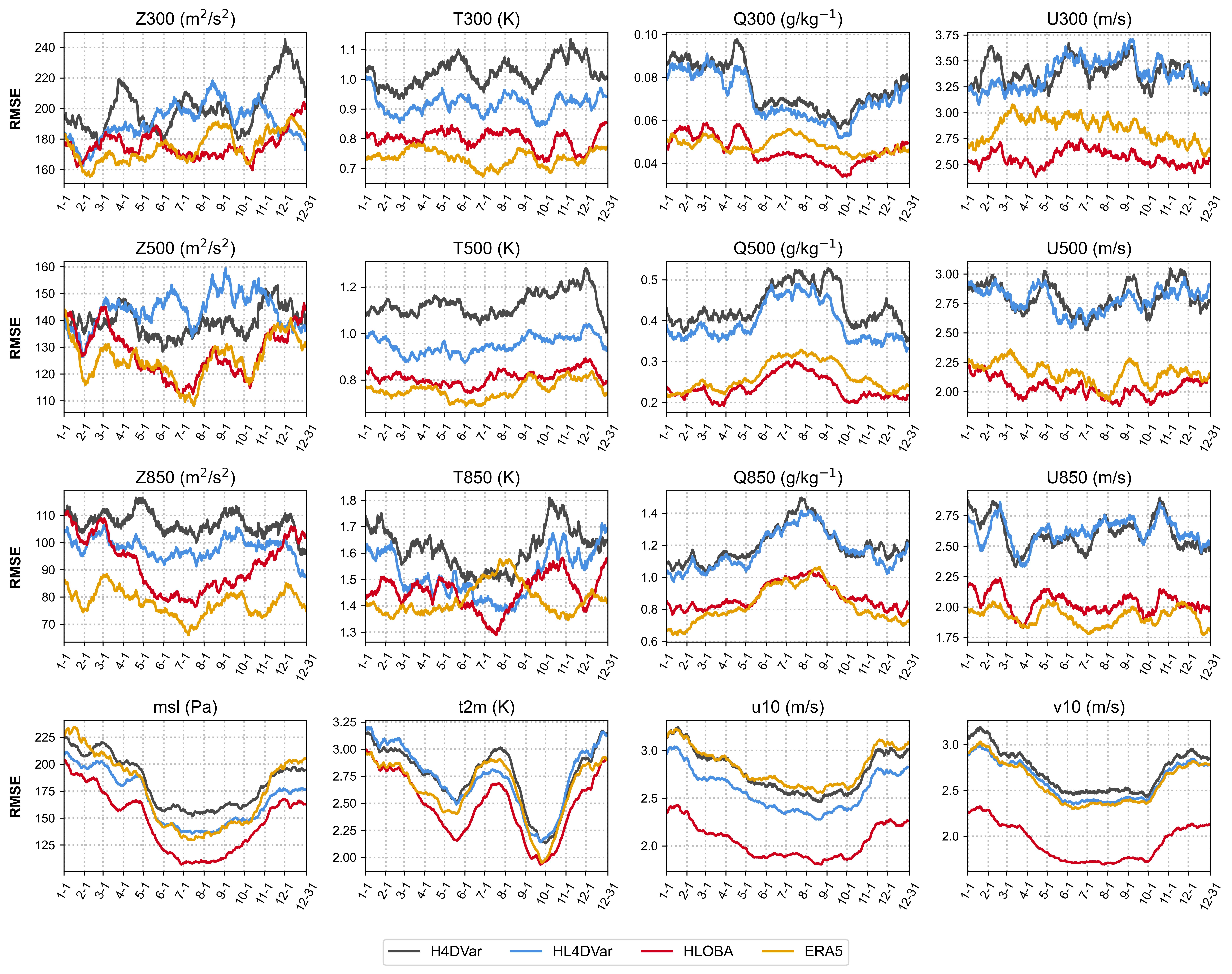}
    \caption{\textbf{Extended analysis performance comparison of HLOBA and other hybrid DA methods with real observations in cycling experiments for 2017.} This figure extends Fig.~\ref{fig:fig2}c by reporting RMSE for variables at 300\,hPa, 500\,hPa, 850\,hPa, and the surfacee, evaluated against observations withheld from assimilation.}
    \label{fig:figS3}
\end{figure}

\begin{figure}[p]
    \centering
    \includegraphics[width=\linewidth]{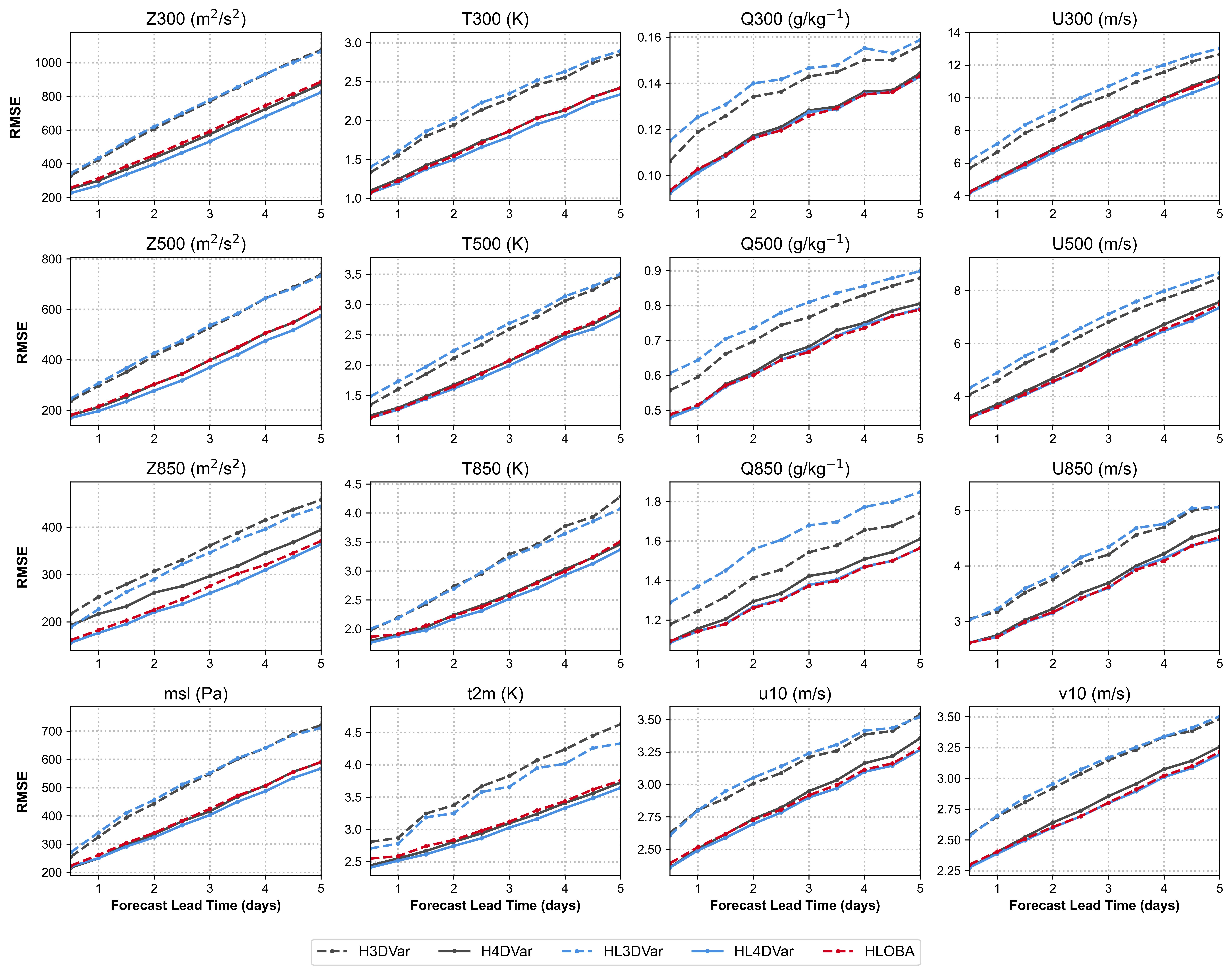}
    \caption{\textbf{Extended forecast performance comparison of HLOBA and other hybrid DA methods with real observations in cycling experiments for 2017.} This figure extends Fig.~\ref{fig:fig2}d by reporting RMSE for variables at 300\,hPa, 500\,hPa, 850\,hPa, and the surfacee, evaluated against all real observations.}
    \label{fig:figS4}
\end{figure}

\begin{figure}[p]
    \centering
    \includegraphics[width=\linewidth]{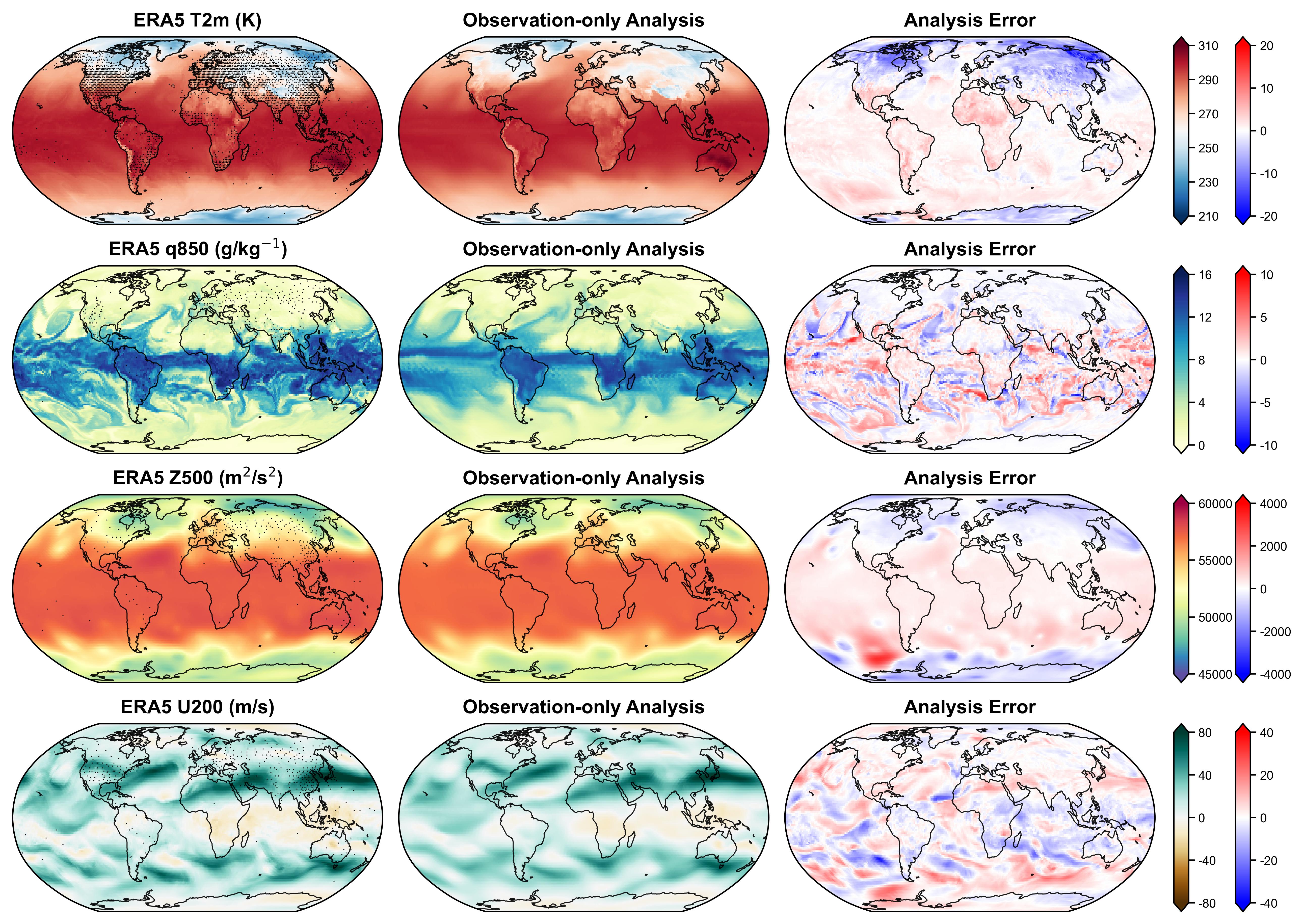}
    \caption{\textbf{The O2Lnet-derived observation-only analysis (OOA) at 0000~UTC on 1 February 2017, using simulated observations sampled from ERA5.} Shown are 2-m temperature (T2m), 850-hPa specific humidity (Q850), 500-hPa geopotential height (Z500), and 200-hPa zonal wind (U200).The observation locations are indicated by black dots. 
    }
    \label{fig:figS5}
\end{figure}

\begin{figure}[p]
    \centering
    \includegraphics[width=\linewidth]{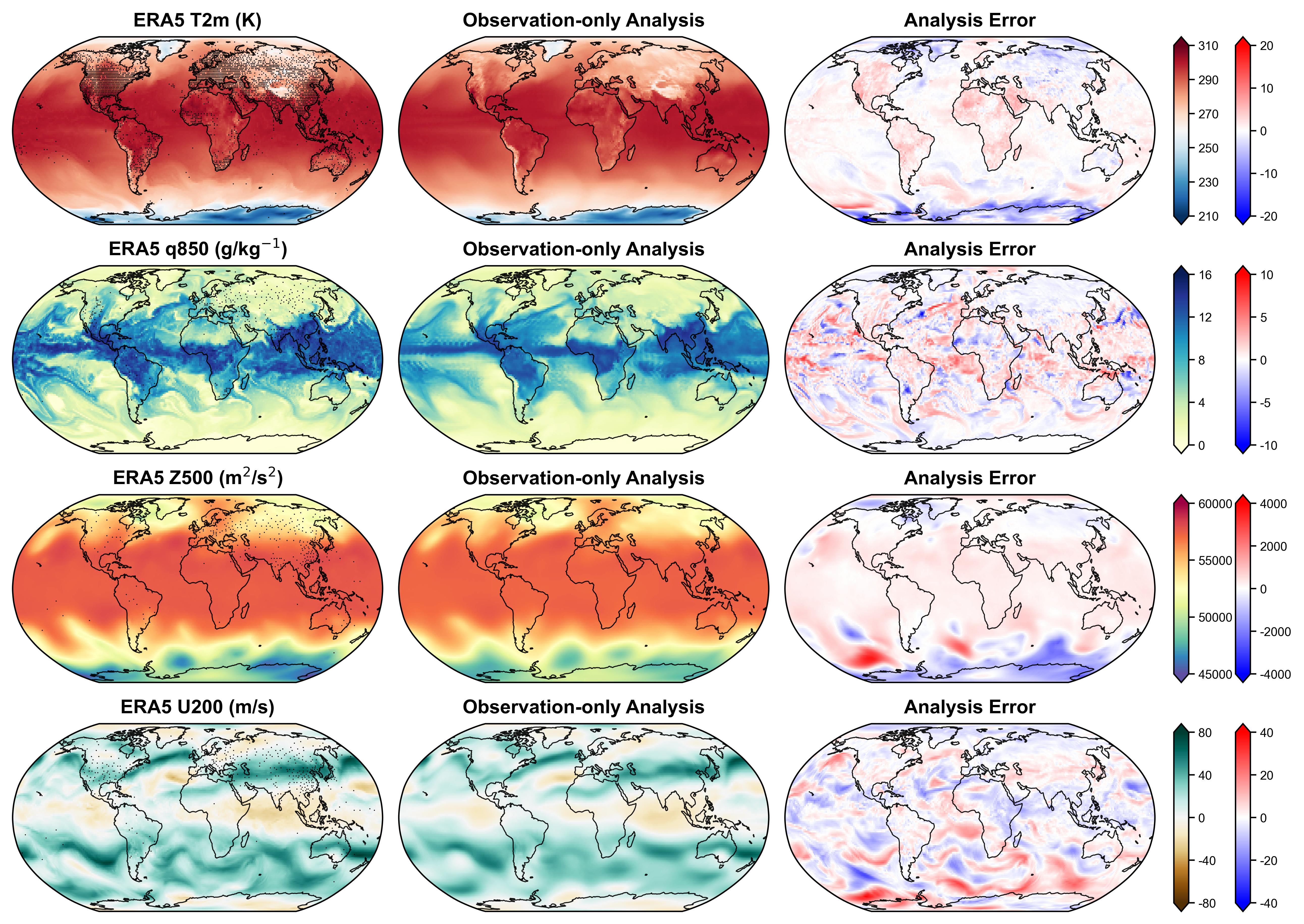}
    \caption{\textbf{Same as Figure~\ref{fig:figS5}, but for the case at 0000~UTC on 1 October 2017.} 
    }
    \label{fig:figS6}
\end{figure}

\begin{figure}[p]
    \centering
    \includegraphics[width=\linewidth]{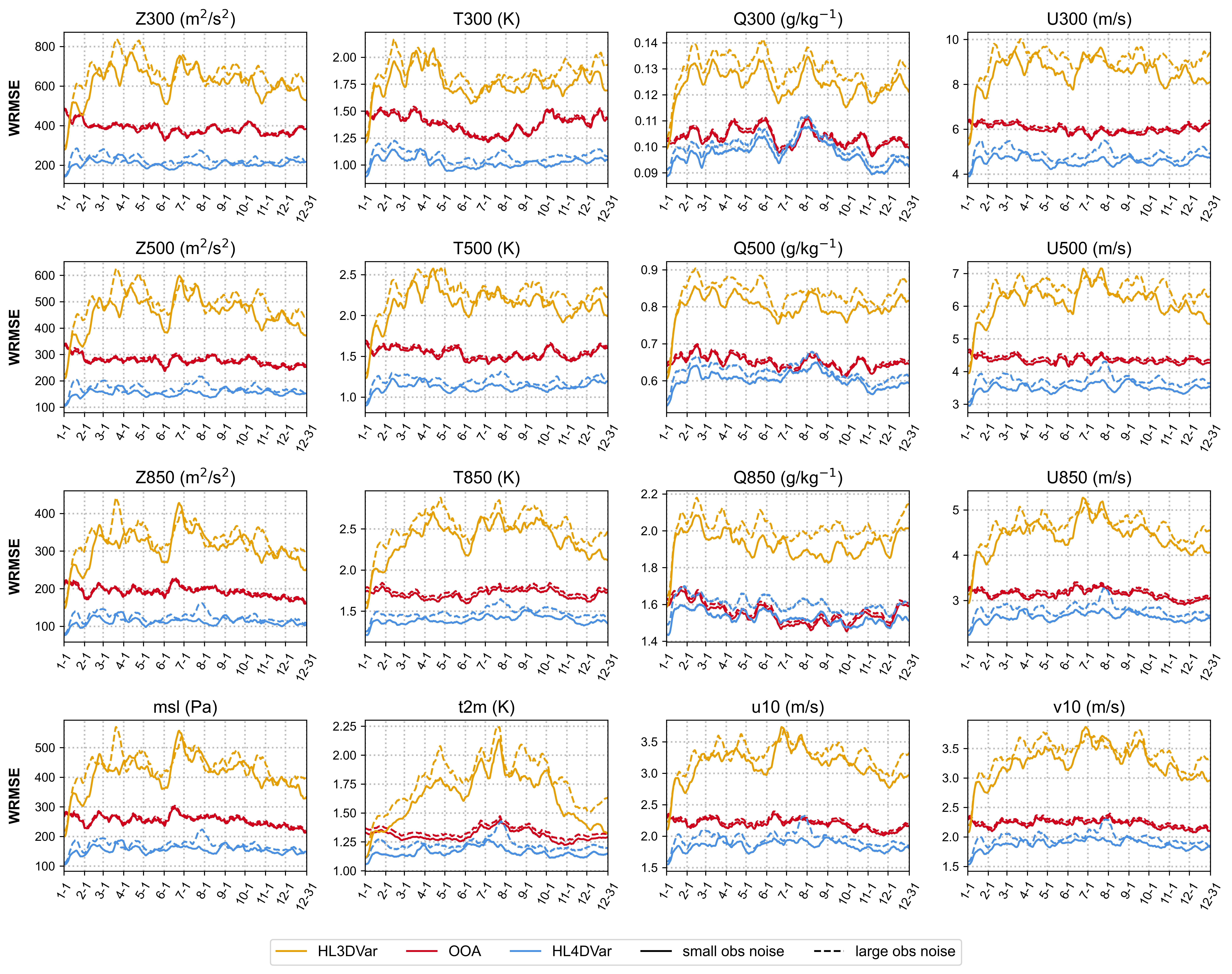}
    \caption{\textbf{Extended accuracy of OOA and its robustness to observation noise in idealized experiments, compared with HL3DVar and HL4DVar} This figure extends Fig.~\ref{fig:fig5}b by reporting WRMSE for variables at 300\,hPa, 500\,hPa, 850\,hPa, and the surface, evaluated against ERA5. 
    }
    \label{fig:figS7}
\end{figure}

\end{document}